\documentclass[12pt]{article}

\RequirePackage{kpfonts}
\RequirePackage[sf,bf,small,raggedright]{titlesec}
\usepackage{dsfont}
\usepackage{comment}
\usepackage{textcomp}
\usepackage{amssymb}
\usepackage{bbm}
\usepackage{fancyhdr}
\usepackage{amsmath}
\usepackage{amsbsy,amsthm}
\usepackage{amscd}
\usepackage{latexsym}
\usepackage{graphicx}   
\usepackage{pdfsync}
\usepackage{blkarray}
\usepackage{multirow}
\usepackage{hyperref}
\usepackage{xcolor}
\usepackage{enumerate}

\usepackage{natbib}
\usepackage{tcolorbox}

\usepackage{tikz}

\usepackage{subcaption}
\usepackage{algorithmic}
\usepackage[justification=centering]{caption}

\usepackage{natbib}
 \bibpunct[, ]{[}{]}{,}{n}{}{,}%

\usepackage{float}
\usepackage{pgf}
\usetikzlibrary{arrows}

\newcommand{\remark}{\medskip\noindent{\bf Remark.} \ }

\newcommand{\field}[1]{\mathbb{#1}}
\newcommand{\E}{\field{E}}

\newcommand{\PROB}{\field{P}}

\newcommand{\IND}[1]{\field{I}_{\{#1\}}}

\newcommand{\defeq}{\stackrel{\rm def}{=}}

\newcommand{\hs}{\widehat{\sigma}}

\newcommand{\Rb}{R_{\alpha}}

\newcommand{\Id}{{\rm Id}}
\newcommand{\de}{{\rm de}}
\newcommand{\pa}{{\rm pa}}

\newtheorem{theorem}{Theorem}
\newtheorem{lemma}[theorem]{Lemma}

\newtheorem{cor}[theorem]{Corollary}

\title{Estimating the history of a random recursive tree\thanks{Simon Briend acknowledges the support of Région Ile de France. Christophe Giraud acknowledges partial support by grant ANR-19-CHIA-0021-01 (BiSCottE, ANR) and by grant ANR-21-CE23-0035 (ASCAI,  ANR and DFG).
G\'abor Lugosi acknowledges the support of Ayudas Fundación BBVA a
Proyectos de Investigación Científica 2021 and
the Spanish Ministry of Economy and Competitiveness grant PID2022-138268NB-I00, financed by MCIN/AEI/10.13039/501100011033,
FSE+MTM2015-67304-P, and FEDER, EU.).
}}

\author{Simon Briend \\
Department of Mathematics, Unidistance,\\
Brig, Switzerland\and
 Christophe Giraud\\
Université Paris Saclay, CNRS,\\
Laboratoire de Mathématiques d'Orsay,\\
91404, Orsay, France\and
G\'{a}bor Lugosi \\
Department of Economics and Business,\\
Pompeu Fabra University, Barcelona, Spain\\
ICREA, Pg. Lluís Companys 23, 08010 Barcelona, \\
Spain Barcelona School of Economics\and
Déborah Sulem\\
Università de la Svizzera Italiana, Switzerland}
\date{}

\begin{document}

\maketitle

\begin{abstract}
This paper studies the problem of estimating
the order of arrival of the vertices in a random recursive tree. Specifically, we study
two fundamental models: the uniform attachment model and the linear preferential attachment model.
We propose an order estimator based on the Jordan centrality measure and define a family of risk measures to quantify the quality of the ordering procedure.
Moreover, we establish a minimax lower bound for this problem, and prove that the proposed estimator
is nearly optimal. Finally, we numerically demonstrate that the proposed estimator outperforms degree-based  and spectral ordering procedures. 
\end{abstract}


\section{Introduction}\label{introduction}

Recursive random trees are often used to model the evolution of relational data over time, with applications ranging from epidemiology to information technology. A rooted labelled tree is said to be recursive when the labels are strictly increasing on every path starting at the root.
When the history of the network is not observed, the task of inferring past states of the network 
is often termed \textit{network archaeology} (\citet{navlkha2011}). Recovering hidden states of the growing network informs on the underlying spreading process and can explain the current network structure.
One of the most studied problems in network archaeology is finding the root of the recursive network, that is, the vertex that first appeared and ``started'' the growing process. This task is closely related to the rumor source detection problem.
It has been extensively analyzed for random recursive tree models, see \citet{DeRe18,LuPe19,Hai70, ShZa11,ShZa16, BuElMoRa17,BuDeLu17, BrDeGo21, JoLo16,JoLo17a, BaBh20, contat2024eve} .

In this paper, we consider the problem of estimating the entire
history of the network, that is, the arrival times of all the vertices
in a random recursive tree. This problem was previously studied
  by \citet{magner2018times} in the case of the preferential
  attachment tree. They conduct their analysis in a setting where the
  position of vertex $1$ is known. This assumption simplifies the
  problem, as will be discussed in this paper. More recently,
  \citet{young2019phase} studied a related problem, where they
  estimate the arrival time of edges in a slightly modified
  preferential attachment graph. They provide numerical experiments
  suggesting a phase transition in the parameter of their model,
  depending on which the history of the graph is recoverable or
  not. In both of these previous works, the authors remark that the
  problem is made hard by the presence of a large set of vertices (or
  edges) that cannot be ordered better than at random. In this paper,
  we introduce a class of measures of the quality of an ordering
  \eqref{eq:risk2} which assign more importance to the error made on
  early vertices. These risk measures define a nontrivial and
  meaningful goal. As was done in \citet{young2019phase}, one may consider this as a question of latent variable estimation. A related statistical problem is the so-called
\emph{seriation}. Seriation is the problem of inferring an 
ordering of points, based on pairwise similarity
or on the adjacency
information between two points. This similarity measure is assumed to statistically decrease with the distance in a latent space and  informs on the latent global order of the points. 
The seriation problem has been studied in various fields, such as in archaeology (\citet{robinson1951method}), bioinformatics (\citet{recanati2017spectral}), and matchmaking (\citet{bradley1952rank}). It has been theoretically analyzed in random graph models such as geometric graphs and graphons (\citet{giraud2023localization,10.1214/21-EJS1940}). In recursive trees, the pairwise affinity between nodes is encoded in the adjacency matrix, and the latent space and latent positions are respectively the temporal line and the arrival times of the vertices. Estimating the temporal order of the vertices in a recursive tree can therefore be interpreted as an instance of the seriation problem.

To estimate the vertices' order, we propose a procedure based on a
centrality measure, specifically on the \emph{Jordan centrality}. We prove
that this procedure is nearly optimal in two random recursive tree
models, namely, the uniform random recursive tree ({\sc urrt}) and the
preferential attachment ({\sc pa}) model. In these models, a tree of
$n \geq 1$ vertices is grown by adding and connecting one vertex at
each time step. To describe the growing process, we assume that the
vertices have intrinsic \emph{labels} from $1$ to $n$. At
each step $t =1,\dots, n$ of the growth, a new vertex, say of label
$j_t$, is picked arbitrarily 
among the set of nodes not yet in the
tree, and added to the tree with the \emph{rank} $t$.  At $t=1$, the
first sampled vertex is the root of the tree. We denote by
$\sigma: \{1,\dots, n\} \to \{1,\dots, n\}$ the ordering (or, ranking)
map of vertices such that $\sigma(j_{t}) = t$. In other words, $\sigma$ is a 
permutation.

The {\sc urrt} and {\sc pa} models differ by the attachment rule used to connect a new vertex at each step $t =2,\dots, n$ of the growth process. In the {\sc urrt} model, the vertex $j_{t}$ is connected by an undirected edge to a vertex sampled uniformly among the vertices 
of the current tree. In the {\sc pa} model, the vertex of the tree is sampled with a probability proportional to its degree. We denote by $T=T_n$ the obtained tree \emph{structure}, that is, the set of nodes with labels in $\{1,\dots, n\}$ and the undirected edges between them. In the statistical problem considered in this paper, after the tree is grown, the rank or arrival time $\sigma(i)$ of each node $i$ is not observed on $T$. The random growing process defines a probability distribution on trees. 
We denote by $\mathbb P$ the corresponding probability distribution and  $\mathbb E$ the associated expectation.   


We note that in these random models, the sampling process is
independent of the labels chosen to identify the vertices, here
$ \{1,\dots, n\}$. Therefore, any coherent ordering procedure should
be \emph{label invariant}, that is, independent of these
labels. Saying that $\hs$ is label invariant means that for any
fixed $T$ and ranking $\sigma$,
\begin{equation}\label{eq:LabelInvariance}
\hs(T,\sigma)\overset{\mathcal{L}}{=}  \hs (T^{\sigma'},\sigma \circ \sigma'),
\end{equation}
for a permutation $\sigma'$, where ${T}^{\sigma'}$ denotes the
tree with label $i$ replaced by $\sigma'(i)$.  Note that the equality
in distribution is a simple equality if the ordering procedure
$\hs$ is deterministic. Let us also remark that an easy way to
transform any ordering procedure into a label invariant ordering
procedure is by applying a random permutation to the labels of the
tree before feeding it to the ordering procedure.

In order to measure the quality of an estimator of the history, we introduce a family of risk measures 
that takes into account the error in the estimated arrival time of each vertex, weighted by a function
of the arrival time. We define the following family of  risk measures
\begin{align}
  \Rb(\hs) \ \defeq \ \E\left[\sum_{i=1}^n \frac{|\hs(i) - \sigma(i)|}{\sigma(i)^{\alpha}}\right] \label{eq:risk2},
\end{align}
where $\alpha > 0$. The expectation is taken with respect to the randomness arising both from the tree and the ordering procedure $\hs$, which can be non-deterministic. The parameter $\alpha$ tunes the importance given
to vertices with small true rank $\sigma(i)$: the higher $\alpha$, the
more weight is given to  vertices with
low rank. Perhaps the most natural choice is $\alpha=1$. 
In that case the risk corresponds to normalizing the
error on the estimation of the arrival time of a vertex by its true
arrival time. We note that it is often the
early stages of a propagation phenomenon that are more relevant,
for example, for designing prevention strategies. Additionally, in
random growing trees, it is harder to accurately order the high-rank
vertices, due to the inherent model symmetries
(\citet{Sreedharan2019InferringTI}).

One way to construct an estimator $\hs$ of the ranking map is to
choose a score function on the set of vertices, and order vertices by
increasing (or decreasing) values. Such a score function could be
based on the likelihood under the tree model. However, the latter is
generally difficult to compute, see \citet{BuDeLu17}. Instead, score
functions based on the degree (\citet{navlkha2011}) or the so-called
rumor centrality (\citet{cantwell2019recovering}) can be computed in
polynomial time. 

Another approach are iterative algorithms that recursively infer
previous states of the tree such as the history sampling algorithms
(\citet{CrXu21, cantwell2019recovering}) and the Peeling procedure
(\citet{Sreedharan2019InferringTI}), which is related to the depth
centrality score. These methods are guaranteed to recover a recursive
ordering of the vertices. Moreover, \citet{CrXu21} show that the
history sampling algorithm outputs confidence sets for the arrival
time of a single vertex with valid frequentist coverage. Besides,
\citet{Sreedharan2019InferringTI} demonstrate that the partial
ordering retrieved by the Peeling procedure has good properties in
settings where the root of the tree can be unambiguously
identified. Nonetheless, there are not yet guarantees on the quality
of the global ordering provided by these methods.

\begin{figure}
\begin{center}
\includegraphics[width=7cm]{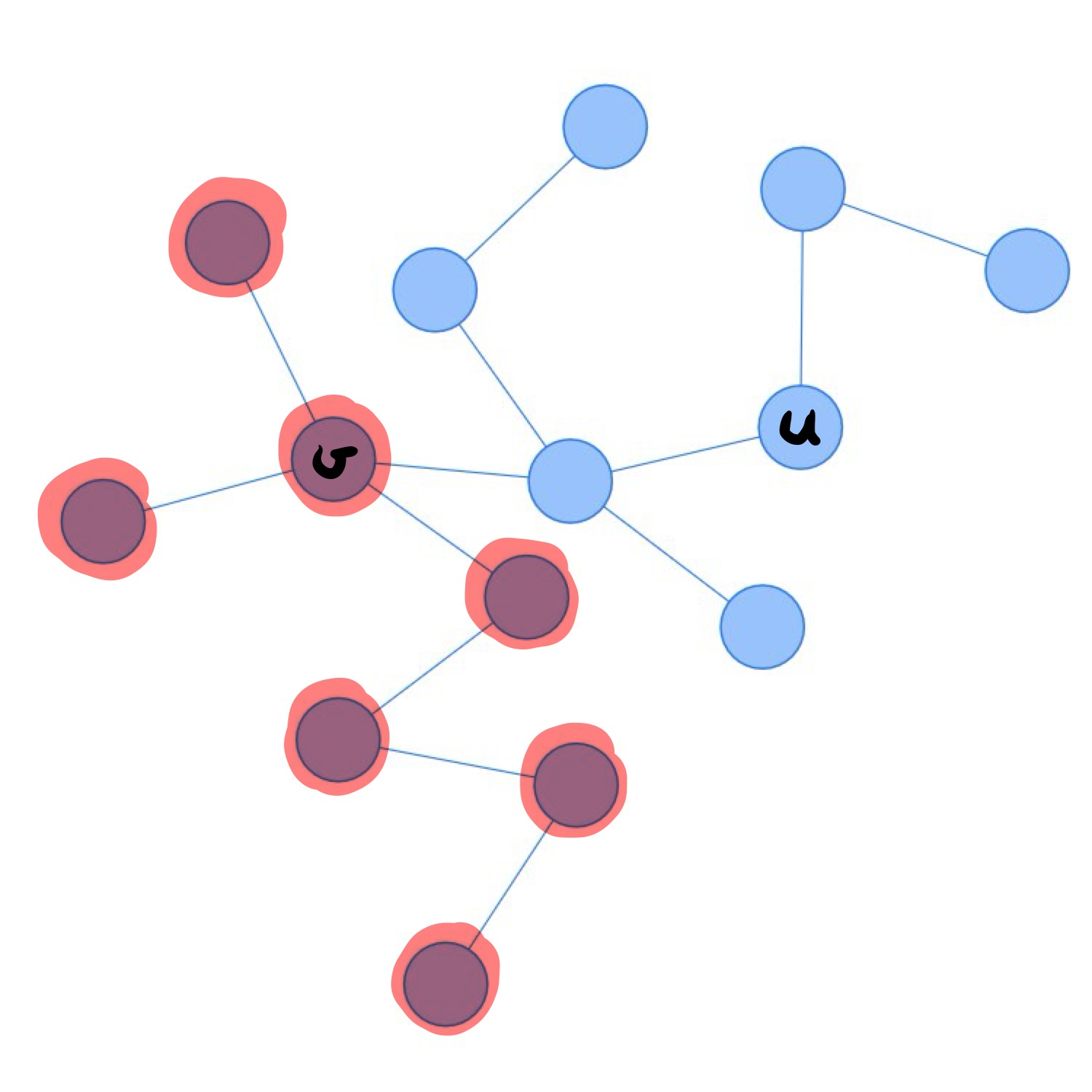}
\caption{An illustration of the subtree $(T,u)_v$, corresponding to nodes highlighted in red.}

\label{fig:HangingSubtree}
\end{center}
\end{figure}

The ordering procedure we propose is based on the Jordan centrality, defined, for a vertex $u \in T$ belonging to a tree $T$, as
\begin{align}\label{eq:jordan}
 \psi_T(u)=\max_{v\in V(T), \ v\sim u} |(T,u)_v|, 
\end{align}
where $(T,u)$ denotes the tree $T$ rooted at $u$, where $u \sim v$ means that $u$ and $v$ are neighbors in $T$, and where $(T,u)_v$ denotes the
subtree of $T$ containing all vertices $w$ such that $v$ lies on
the path connecting $w$ to $u$ (see Figure \ref{fig:HangingSubtree}). 
Somewhat informally, we call $(T,u)_v$
the subtree hanging from $v$ in the rooted tree $(T,u)$. The maximum in \eqref{eq:jordan} is taken over vertices $v$ of the tree
that are connected to vertex
$u$ by an edge. Intuitively, if a vertex is \emph{central}, then none of the
subtrees hanging from it can be too large. Therefore, the lower
$\psi_T(i)$, the more central is vertex $i$. It is
straightforward to see that $(\psi_T(u))_{u \in T}$ only depends on
the structure of the tree and not on the labels of its vertices. We
then define $\hs_{J} : \{1, \dots, n\} \to \{1, \dots, n\}$ the
ordering obtained by ranking the vertices by increasing value of
Jordan centrality--breaking ties at random. This estimator is label
invariant. An equivalent formulation of this algorithm is to estimate
the position of vertex $1$ by the Jordan centroid, rooting the tree at
this vertex and then ordering vertices by the size of their hanging
subtree in the rooted tree. Thus, if the exact position
of vertex $1$ was known, we would be ordering vertices by the number of their
descendants. 

While the risk defined in \eqref{eq:risk2} can be computed for any
value $\alpha > 0$, we restrict $\alpha$ to a range of values which
are relevant for our ordering problem. Specifically, we only consider
$\alpha \geq 1$, since for $\alpha<1$ the problem becomes trivial, since even a random permutation has a risk which is minimax optimal up to constant factor (see
Appendix \ref{app:alpha}). In Theorems \ref{prop:urrtLowerBound}
and \ref{prop:PALowerBound}, we provide minimax lower
bounds for the risk $\Rb(\hs)$ in the {\sc urrt} and {\sc pa} model,
for any label-invariant estimator $\hs$. Then, in Theorems
\ref{prop:urrtErrorUpperBound} and
\ref{prop:PAErrorUpperBound} upper bounds for the risk of the Jordan
ordering are obtained. Finally, in Corollaries
\ref{cor:urrtErrorUpperBound} and \ref{cor:PAErrorUpperBound} we prove
that our proposed estimator is minimax 
optimal up to constant factors,
in a non-trivial range of parameters $\alpha$. In the following table,
we summarise our findings. For $\alpha\geq 1$, we denote by $R^*_{\alpha}$ the optimal
risk, and $\Rb(\hs_{J})$ the risk of the Jordan ordering.

\begin{center}
\begin{tabular}{c c c } 
 \hline
  &   {\sc urrt} & {\sc pa}   \\ 
 \hline 
 
  $\Rb^{*}$  & $ \geq n^{2-\alpha}/70 \vee 1/2 $ & $
                                                                   \geq
                                                                   n^{2-\alpha}/70
                                                                   \vee 1/2 $ \\
  
  \hline
 $\Rb(\hs_{J})$   & $ = \mathcal{O} \left(n^{2-\alpha}+\log^4(n)\right)  $ & $ = \mathcal{O}\left( n^{2-\alpha}+n^{3/4}\right)  $   \\ 
 
\hline

\end{tabular}
\end{center}

We also compare numerically the performance of the Jordan estimator
with other ordering procedures  in a simulation study. 

In the rest of this section, we review previous works and
introduce some notation. Then, in Section \ref{sec:urrt}, we analyze
the Jordan ordering in the {\sc urrt} model. Next, we consider the
{\sc pa} model and prove analogous results in Section
\ref{sec:PA}. Finally, in the Supplementary Material, we report the
results of our simulation study and compare the empirical performance
of the Jordan estimator to alternative methods based on the degree
centrality, a peeling method (\citet[Section 2.3]{navlkha2011}) and a
spectral method commonly used in seriation problems
(\citet{recanati2018reconstructing}).

\subsection{Related work}

Most methods for ranking the vertices of a random recursive tree have
been introduced for the root-finding problem, that is,
recovering a vertex (or a set of vertices) that is (contains) the
root. For this problem, maximum likelihood estimators
(\citet{BuDeLu17,Hai70, BrDeGo21}) and estimators based on rumor centrality
(\citet{ShZa11,ShZa16}) have been proposed and analyzed. Jordan
centrality is another measure of centrality
used by \citet{BuElMoRa17,BuDeLu17} to construct
confidence sets. \citet{JoLo16,JoLo17a, BaBh20} study the
persistence of the most central nodes in random recursive trees.
Furthermore, while the vertex with maximum degree is generally not a
good estimator of the root in the {\sc urrt} model, in the {\sc pa}
model 
pairs degree centrality is useful for retrieving the first vertex (\citet{BaBh21,contat2024eve}). Some recent work studies root-finding in Galton-Walton trees (\citet{BrDeGo21}) and more general graphs  (\citet{CrXu21a, briend2023archaeology}).

\citet{CrXu21} propose a general history-sampling procedure for
network archaeology, which can be applied to the problem of estimating arrival times.
The history sampling algorithm outputs a confidence set of rankings that
contains the true one with high probability. However, there is no
known bound of the size nor the average global error of an ordering in
this confidence set.

The vertex arrival-time estimation problem bears some similarity to
the seriation problem, though in the former, the dependence
is intrinsically related to the tree structure. For example, in a
random geometric graph (\citet{gilbert1961random}), the seriation 
problem is to estimate the position of the random points. 
Since there is no time structure
in seriation, different metrics for the error are used, such as 
the maximum distance between the true and estimated latent
position. Examples of methods are provided by
\citet{giraud2023localization}. 
Another widely studied seriation method
consists in ordering latent points by a spectral method on the graph
Laplacian (see Supplementary material and
\citet{recanati2018reconstructing} for details). They give guarantees
for the quality of their method when the observed adjacency matrix is a
perturbed Robinson matrix. 
The expected adjacency matrices of both {\sc urrt} and {\sc pa} trees are Robinson, and therefore 
 the models studied here can be viewed as perturbed Robinson matrices. Nonetheless, none of the above-mentioned
papers gives any insight about seriation in {\sc urrt} and {\sc pa} trees.

\subsection{Notation}\label{sec:notation}

Let $\pi_n$ be the set of permutations of $[n] := \{1,2, \dots, n\}$,
and let $\mathbb{T}_n$ be the set of un-labelled trees of size $n$. We
denote by $\text{{\sc urrt}}(n)$ the distribution of a tree
$\mathcal{T}_{n}$ of size $n \geq 1$, generated from the uniform
attachment model.
Similarly, we
denote by $\text{{\sc pa}}(n)$ the distribution of a tree sampled from the
preferential attachment model. Moreover, we decompose the tree as
$\mathcal{T}_{n} = (T_{n}, \sigma_{n})$, where $ T_{n} $ is the shape
of the tree and $\sigma_{n}$ is the recursive ordering of the vertices
in $\mathcal{T}_{n}$. For simplicity, we drop the subscript $n$ when
the size of the tree is fixed and clear from the context. We denote by
$\mathbb P$ the probability distribution under the tree growing
process and $\mathbb E$ the corresponding expectation. 

Recall that for a tree $T$
and a vertex $u \in T$, we denote by $(T,u)$ the tree rooted at
$u$. For a rooted tree $(T,u)$ and a vertex $v$ we denote by
$(T,u)_{v}$ the subtree of $T$ consisting of all vertices $w$ such
that $v$ lies on the path connecting $w$ to $u$ (see Figure \ref{fig:HangingSubtree}). For two vertices
$u, v \in T$, $u \sim v$ means that $u$ is a neighbor of $v$ in $T$
(and reciprocally). In a rooted tree $(T,u)$, we say that $w$ is a
child of $v$ if $w$ is in $(T,u)_v$. We denote by $\de_n(u)=|(\mathcal{T}_n,1)_u|-1$ the number
of descendants of $u$ in $\mathcal{T}_n$.
For simplicity, we drop the subscript $n$ and
use $\de(u)$ when the size of the tree is fixed and clear from the
context.


Recall the definition of the Jordan centrality; for a tree $T$ and vertex $u \in T$,
\begin{align}
 \psi_T(u)=\max_{v\in T, v \sim u} |(T,u)_v|. 
\end{align}
We denote by $c$ a centroid of $T$, defined as
$c = \arg \min_{u \in T} \psi_T(u)$. It is well-known that any tree
has at least one and at most two centroids. Moreover, for a vertex
$u \in T$ that is not a centroid, the subtree $(T,u)_v, \: v \sim u$
with maximum size, contains all centroids.

The Jordan ordering procedure consists in ordering points by
increasing values of $\psi$ (ties being broken randomly). Equivalently,
it consists in 
rooting the tree at $c$ and order vertices by $|(T,c)|_u$. We 
use $\hs_J$ to refer to the Jordan ordering of $T_n$. As
noted in the introduction, the Jordan centrality does not depend on
the labelling of the tree (only on its shape), and so is a label
invariant ordering.

\section{The uniform attachment model}\label{sec:urrt}

In this section, we focus on the uniform attachment model as the
random growing process of the tree. We first present a
lower bound for the risk $\Rb(\hs)$ of any label-invariant estimator $\hs$
of the vertices order. 

\subsection{A lower bound}\label{sec:urrtLowerBound}

In the next proposition, we provide a lower bound for the risk $\Rb(\hs)$ for any label-invariant estimator of the recursive ordering in the {\sc urrt} model. Define, for any $n \geq 1$, the optimal risk by
$$ R^*_{\alpha}:=\min_{\hs \in \Pi_n} \Rb(\hs), $$
where $\Pi_n$ is the set of label-invariant recursive orderings.

\begin{theorem}\label{prop:urrtLowerBound}
In the {\sc urrt} model, we have, for all $\alpha  >0$ and $n\geq 200$,
$$ R^*_{\alpha}\geq \frac{n^{2-\alpha}}{70}.  $$
\end{theorem}

\begin{proof} For a tree $T$ and an ordering of its vertices $\sigma$, let $\tau=\sigma^{-1}$ (i.e, $\tau(i)$ is the label of the vertex that arrives at time $i$). We start by recalling that
$$ \Rb(\hs)=\E\left[ \sum_{j=1}^n \frac{\left|\hs(j)-\sigma(j)\right|}{\sigma(j)^{\alpha}}  \right]=\E\left[ \sum_{j=1}^n \frac{\left|\hs\circ\tau(j)-j\right|}{j^{\alpha}}  \right],$$
which is lower bounded as follows

\begin{align}\label{eq:RbSimpleLowerBound}
 \Rb(\hs) &\geq \sum_{j=\left\lfloor n/2 \right\rfloor +1}^{\left\lfloor 3n/4 \right\rfloor} \E\left[\frac{\left|\hs\circ\tau(j)-j\right|}{j^{\alpha}}\right]+\sum_{j=\left\lfloor 3n/4 \right\rfloor +1}^{n} \E\left[\frac{\left|\hs\circ\tau(j)-j\right|}{j^{\alpha}}\right]\\
 & \geq \frac{1}{n^{\alpha}}  \sum_{j=\left\lfloor n/2 \right\rfloor +1}^{\left\lfloor 3n/4 \right\rfloor} \E\left[ \big|\hs\circ\tau(j)-j\big|+\bigg| \hs\circ\tau\left(\left\lfloor \frac{n}{4} \right\rfloor+j\right) -\ \left\lfloor \frac{n}{4} \right\rfloor-j\bigg| \right] .
\end{align}
The problem is reduced to a control of each term of the summand. For a labelled tree $T$ and a permutation $\gamma$, we denote by $T^{\gamma}$ the tree with $\gamma$ applied to its labels. For $ \lfloor n/2 \rfloor+1\leq j \leq \lfloor 3n/4\rfloor$, fix $\gamma=\left(\tau(j),\tau(\left\lfloor n/4 \right\rfloor +j) \right)$, that is, the permutation sending $j$ to $\left\lfloor n/4 \right\rfloor +j$ and vice versa, while keeping all other elements of $[n]$ in place. Introduce the event

$$ \Omega_j:=\left\{ \tau(j) \text{ and } \tau(\left\lfloor n/4\right\rfloor +j) \text{ are leaves, connected to vertices of rank  } \le n/2 \right\} . $$
First, we check that $\Omega_j$ is an event whose probability is
bounded away from $0$. We note that for $\Omega_j$ to occur, it
suffices that
\begin{itemize}
\item vertex $\tau(j)$ connects to a vertex of rank at most $n/2$. This happens with probability $\left\lfloor n/2 \right\rfloor /(j-1) $. 
\item For times ranging from $j+1$ to $\lfloor n/4 \rfloor +j-1$ new vertices connect to vertices different from $\tau(j)$. This happens with probability 
$$ \prod_{k=j+1}^{\left\lfloor n/4 \right\rfloor+j-1} \frac{k-2}{k-1}.$$
\item Vertex $\tau(\lfloor n/4 \rfloor +j)$ connects to a vertex of rank at most $n/2$. This happens with probability $\left\lfloor n/2 \right\rfloor /(\lfloor n/4\rfloor+j-1) $. 
\item For times ranging from $\lfloor n/4 \rfloor +j+1$ to $n$ new vertices connect to vertices different from $\tau(j)$ and $\tau(\lfloor n/4 \rfloor +j)$. This happens with probability 
$$\prod_{k=\left\lfloor n/4 \right\rfloor+j+1}^{n} \frac{k-3}{k-1}.$$
\end{itemize}
Finally, note that from the definition of the {\sc urrt} model, the four events corresponding to the four items above are independent. Thus
\begin{align*}
\PROB\left\{ \Omega_j \right\}& = \frac{\left\lfloor n/2 \right\rfloor }{j-1} \cdot \frac{\left\lfloor n/2 \right\rfloor }{\left\lfloor n/4 \right\rfloor +j-1} \cdot \prod_{k=j+1}^{\left\lfloor n/4 \right\rfloor+j-1} \frac{k-2}{k-1} \cdot \prod_{k=\left\lfloor n/4 \right\rfloor+j+1}^{n} \frac{k-3}{k-1} \\
&= \frac{\left\lfloor n/2 \right\rfloor }{j-1} \cdot  \frac{\left\lfloor n/2 \right\rfloor }{\left\lfloor n/4 \right\rfloor +j-1} \cdot \frac{j-1}{\left\lfloor n/4 \right\rfloor+j-2} \cdot \frac{(\left\lfloor n/4 \right\rfloor+j-2)(\left\lfloor n/4 \right\rfloor+j-1)}{(n-2)(n-1)},
\end{align*}
which simplifies to
\begin{equation}\label{eq:OmegaPositiveProba}
 \PROB\left\{ \Omega_j \right\} \geq \frac{1}{4} \left(1-\frac{1}{n-1}\right) .
\end{equation}
 The first step is to use \eqref{eq:OmegaPositiveProba} to control one of the summands in \eqref{eq:RbSimpleLowerBound} by conditioning on $\Omega_j$.
\begin{eqnarray*}
\lefteqn{
 \E\left[ \left|\hs\circ\tau(j)-j\right|+|\hs\circ\tau\left(\left\lfloor n/4 \right\rfloor+j\right) -\left(\left\lfloor n/4 \right\rfloor+j\right)| \right] } \\ 
& \geq & \frac{1}{4} \left(1-\frac{1}{n-1}\right) \E\left[ \left. \big|\hs\circ\tau(j)-j\big|+\bigg| \hs\circ\tau\left(\left\lfloor \frac{n}{4} \right\rfloor+j\right) -\left(\left\lfloor \frac{n}{4} \right\rfloor+j\right)\bigg|  \ \right\vert \ \Omega_j \right].
\end{eqnarray*}
We then decompose on each possible realization of a recursive tree

\begin{align*}
&\E\left[ \left. \big|\hs\circ\tau(j)-j\big|+\bigg| \hs\circ\tau\left(\left\lfloor \frac{n}{4} \right\rfloor+j\right) -\left(\left\lfloor \frac{n}{4} \right\rfloor+j\right)\bigg|  \ \right\vert \ \Omega_j \right] \\
&=\sum_{t\in\mathbb{T}}\PROB\left\{ T=t\mid \Omega_j \right\}  \E\left[ |\hs\circ\tau(j)-j|  \ \mid \ \Omega_j, \ T=t \right] \\
& \ \  \ \  + \sum_{t\in\mathbb{T}}\PROB\left\{ T=t^{\gamma}\mid \Omega_j \right\}  \E\left[ \left. \bigg| \hs\circ\tau\left(\left\lfloor \frac{n}{4} \right\rfloor+j\right) -\left(\left\lfloor \frac{n}{4} \right\rfloor+j\right)\bigg|  \ \right\vert \ \Omega_j, \ T=t^{\gamma} \right],
\end{align*}
which is a valid decomposition since $t\mapsto t^{\gamma}$ is a bijection from $\mathbb{T}$ to itself. Theorem 4 of \citet{CrXu21} states that, in the {\sc urrt} model, two trees having the same shape but different recursive orders have the same probability. Since on the event $\Omega_j$, $t$ is recursive if and only if $t^{\gamma}$ is recursive, then 

$$ \PROB\left\{ T= t\mid \Omega_j \right\}=\PROB\left\{ T=t^{\gamma}\mid \Omega_j \right\}. $$
As a consequence, the above expression factorizes to 
\begin{align}\label{eq:TreeDecomposition}
&\E\left[ \left. |\hs \circ\tau(j)-j|+\left| \hs \circ\tau\left(\left\lfloor \frac{n}{4} \right\rfloor+j\right) -\left\lfloor \frac{n}{4} \right\rfloor-j\right|  \ \right\vert \ \Omega_{j} \right] =\sum_{t\in\mathbb{T}}\PROB\left\{ T=t\mid \Omega_{j} \right\} \times \nonumber \\
&  \left(\E\left[ |\hs\circ\tau(j)-j|  \ \mid \ \Omega_j, \ T=t \right]+\E\left[\left. \left| \hs\circ\tau\left(\left\lfloor \frac{n}{4} \right\rfloor+j\right) -\left(\left\lfloor \frac{n}{4} \right\rfloor+j\right)\right|  \ \right\vert \ \Omega_j, \ T=t^{\gamma} \right] \right).
\end{align}
\noindent The label invariant condition implies that
$$ \hs[T^{\gamma}]\circ\gamma \overset{\mathcal{L}}{=} \hs[T], $$
and in particular,
$$ \left( \hs\circ \tau(j) \ \mid \Omega_j,T=t\right)\overset{\mathcal{L}}{=} \left(\hs\circ\tau\left( \left\lfloor n/4 \right\rfloor +j \right) \ \mid \Omega_j,T=t^{\gamma}\right),$$
\noindent which directly implies that 

\begin{multline*}
\E\left[ |\hs\circ\tau(j)-j|  \ \mid \ \Omega_j, \ T=t \right]
 + \E\left[\left. \left| \hs\circ\tau\left(\left\lfloor \frac{n}{4} \right\rfloor+j\right) -\left\lfloor \frac{n}{4} \right\rfloor-j\right|  \ \right\vert \ \Omega_j, \ T=t^{\gamma} \right] \geq \lfloor n/4\rfloor  .
\end{multline*}
\noindent By plugging the above inequality in \eqref{eq:TreeDecomposition}
$$\E\left[ |\hs\circ\tau(j)-j|+\left| \hs\circ\tau\left(\left\lfloor \frac{n}{4} \right\rfloor+j\right) -\left\lfloor \frac{n}{4} \right\rfloor-j\right| \right] \geq \frac{n-4}{16} \left( 1-\frac{1}{n-1}\right).$$
\noindent Now, plugging the above inequality in \eqref{eq:RbSimpleLowerBound} yields
$$\Rb(\hs)\geq \frac{n^{2-\alpha}}{70},$$
for all $n\geq200$.
\end{proof}

\remark
In the {\sc urrt}, vertices $1$ and $2$ are indistinguishable. Indeed, when the tree has size $2$, vertices $1$ and $2$ have exactly the same properties. Thus, no label invariant ordering procedure can assign order $1$ to vertex $1$ with probability higher than $1/2$. As a result, we obtain, for any $\alpha$, the trivial lower bound

$$ R_{\alpha}^*\geq \frac{1}{2},$$
which improves the bound of Theorem \ref{prop:urrtLowerBound} for $\alpha \ge 2$, and therefore 
 Theorem \ref{prop:urrtLowerBound} is non-trivial when $\alpha <2$.

\subsection{An auxiliary ``descendant-ordering'' procedure}
\label{sec:proxyurrt}

In the sequel, we estabish upper bounds for the risk of Jordan ordering.
Since this is a label-invariant procedure, we may assume, 
without loss of generality, that $\sigma=\Id$ is the 
identity permutation.
In other words, the arrival time of a vertex and its label are the
same. When the context is clear, 
vertex labels and arrival times are used interchangeably.

In order to analyze the Jordan ordering, we introduce an auxiliary
centrality measure and the corresponding estimator of vertex arrival times. 
As observed in the
introduction, the Jordan ordering procedure consists in estimating the
position of vertex $1$ by the Jordan centroid $c$ and ordering vertices
according to the values of $|(T,c)_u|$. If $c$ was replaced by 
vertex $1$, this measure minus $1$ would correspond to the number of descendants of $u$. 
Thus, a
natural ordering is to order vertices by the number of their
descendants, that is, the ordering according to the values of $|(T,1)_u|-1$ (i.e., by the value of $|(T,1)_u|$). 
We call this \emph{descendant ordering}, noting that, as before, ties
are broken at random.
Note that descendant ordering is not a valid procedure, since the location of the 
root vertex is not known. On the other hand, the number of descendants is
easily analyzed by P\'olya urns, and our approach is based on comparing
Jordan ordering to this auxiliary procedure. 
In this section we prove an upper bound for difference of the risk of
both procedures.  For a tree $T$ and for each $u \in T$, we define the descendant centrality
$$ \psi'_T(u)=n+1-\de(u), $$
where $\de(u)=|(T,1)_u|-1$ is the number of descendants of $u$, as defined in Section \ref{sec:notation}.
We denote by $\hs'$ the ordering of the vertices induced by sorting the values of $ \psi'_T$ in increasing order. 

In the following lemma, we first prove that for the {\sc urrt} model, the Jordan centrality $\psi_{T}$, defined in \eqref{eq:jordan}, and the descendant centrality $\psi_{T}'$ coincide for most vertices. Furthermore, we prove bounds on both the number of nodes for which $\psi_{T}$ may differ from $\psi_{T}'$ and the estimated rank of vertex $1$. We recall that $1$ and $c$ denote respectively the root and the rank of a centroid of the tree.

\begin{lemma}\label{lem:JordanVsHanging}
Let $T \sim \text{{\sc urrt}}$, let $c\in[n]$ be a centroid of $T$ and let $\{1\to c\}$ be the set of vertices on the path connecting $1$ to $c$ in $T$. Then
\begin{itemize}
\item for any $v \in [n] \backslash \{1\to c \}$, we have
\begin{align*}
\psi_T(v)=\psi_T'(v);
\end{align*}

\item  there exists a universal constant $K$ such that $c$ is stochastically dominated by an exponential random variable with mean $K$;

\item  for $\epsilon\leq0.2$, with probability at least $1-5\epsilon$
$$ \hs_J(1) \leq 2.5\frac{\log(1/\epsilon)}{\epsilon}. $$
\end{itemize}
\end{lemma}

\begin{figure}
\begin{center}
\includegraphics[scale=0.16]{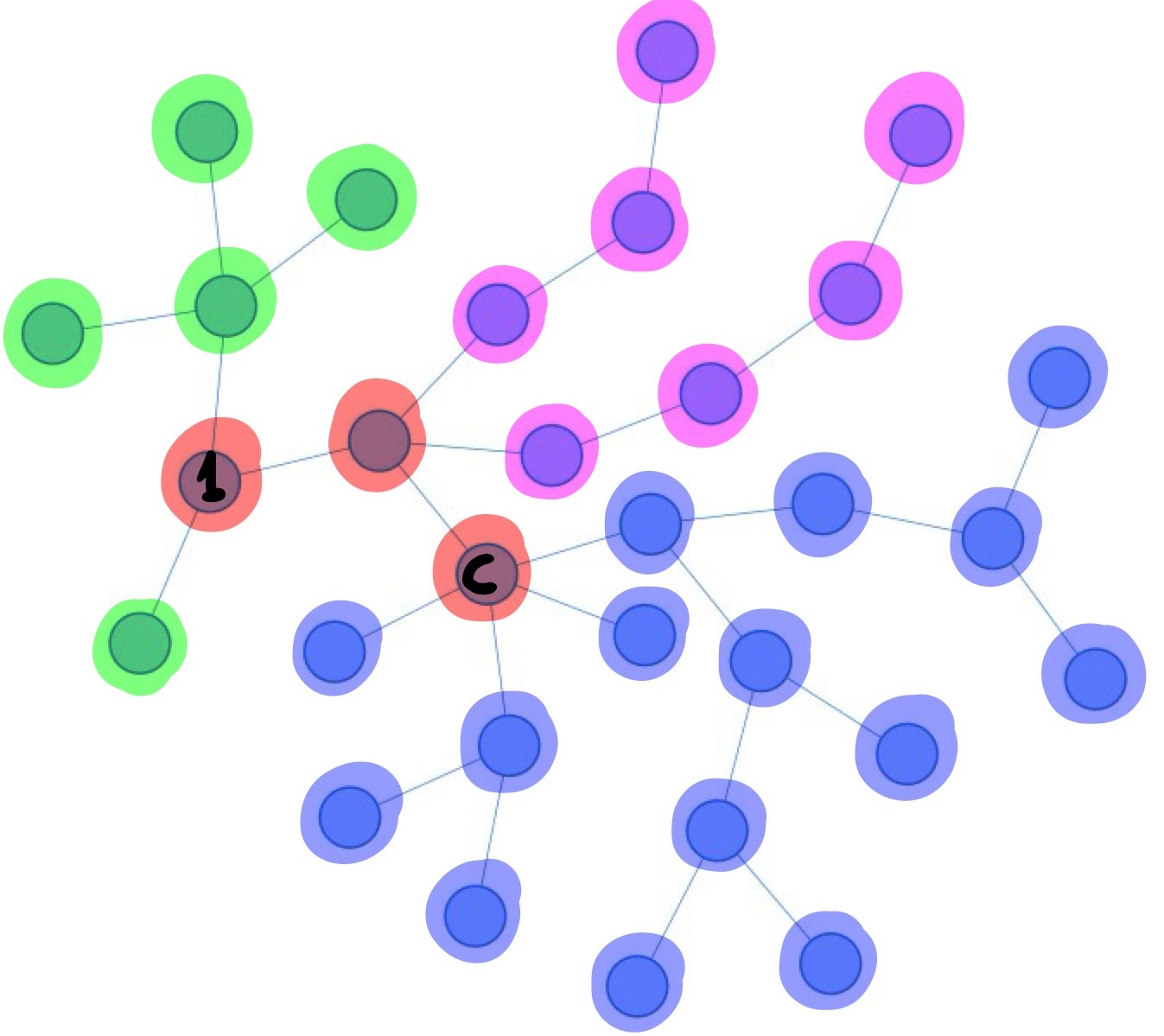}
\caption{Sketch of a tree and its centroid. Circled in red are the vertices of the path $\{1\to c\}$ (case $1$). Blue vertices correspond to case $2$, green to case $3$ and purple vertices to case $4$.}
\label{fig:JordanVsHanging}
\end{center}
\end{figure}

\begin{proof}
  Let $T \sim \text{{\sc urrt}}$. First, we decompose the vertices of
  $T$ in four sets as shown in 
 Figure \ref{fig:JordanVsHanging}: 
case
  $1$ corresponds to the set $ \{ 1 \to c \}$ of nodes connecting the
  root to the centroid, case $2$ to the vertices of
  $(T,1)_c\setminus \{c\}$, case $3$ to the vertices of
  $(T,c)_1\setminus\{1\}$ and finally case $4$ to the vertices of
  $(T,1)_i\setminus\{i\}$ for $i\in \{ 1 \to c \}\setminus\{1,c\}$. As
  we mentioned before, it is well known that for a non-centroid vertex
  $u$, its neighbor maximizing $|(T,u)_v|$ is such that $|(T,u)_v|$
  contains any centroid. Note that for each vertex $u$ in cases $2$,
  $3$ and $4$, for $v\sim u$ such that the subtree $(T,u)_v$ contains
  $c$, $(T,u)_v$ also contains vertex $1$. As a consequence,
  $\psi_T(u)=|(T,u)_{\pa(u)}|$, where $\pa(u)$ is the ``parent'' of
  $u$. But by definition,
  $|(T,u)_{\pa(u)}| = n+1 - \de(u) = \psi_{T}'(u)$, concluding the proof
  of the first part of the lemma.

  \citet{moon2002centroid} showed that the rank of the centroid $c$ is
  dominated by an exponential random variable of mean $K$, for a
  universal constant. The third statement follows from Theorem $3$ of
  \citet{BuDeLu17}.
\end{proof}

In the following lemma, we bound the risk of $\hs_J$ by that of the
descendant ordering $\hs'$. To compare the risks of the Jordan
  and descendant ordering, we introduce a simple coupling between the
  two of them, so that ties are broken at random in the same way in
  both orderings. In a tree $T$ of size $n$, first compute the Jordan
  ordering (where ties are broken uniformly at random). Then, to
  compute the descendant ordering, one must choose how to break
  ties. Let us fix an integer $k$. We denote by $v_1,\ldots,v_j$ the
  vertices of $T$ having exactly $k$ descendants. From Lemma
  \ref{lem:JordanVsHanging} we know that vertices in
  $\{v_1,\cdots,v_j\}\cap[n]\backslash \{1\to c\}$ have the same Jordan centrality. These ties have been broken uniformly at random in the Jordan ordering. For $j'=|\{v_1,\ldots,v_j\}\cap[n]\backslash \{1\to c\}|$, without loss of generality, we can assume that $\{v_1,\ldots,v_j\}\cap[n]\backslash \{1\to c\}=\{v_1,\ldots, v_{j'}\}$ and that $\hs(v_1)<\cdots<\hs(v_{j'})$. In the descendant ordering we break the ties of $\hs'(v_1),\ldots,\hs'(v_j)$ uniformly at random conditioned on $\hs'(\sigma_1)<\cdots<\hs'(\sigma_{j'})$. By doing so for every $k$, the ties of the descendant ordering are broken uniformly at random but in the same manner as in the Jordan ordering. This coupling allows us to compare their respective risks.

\begin{lemma}\label{prop:ProxyIsGood}
Let  $T \sim  \text{{\sc urrt}}$. For $\alpha >0$

$$ \Rb(\hs_J) \leq \Rb(\hs') +K\sum_{i=1}^n\frac{1}{i^{\alpha}}+C\log^4(n) , $$
where $C>0$ is a constant (not depending on $\alpha$).
\end{lemma}

\begin{proof}
Recall that $\sigma = \Id$, that is, we use the same integer to denote the label of a vertex and its arrival time. We first decompose the global risk $\Rb(\hs_J)$ into 
\begin{align*}
 \Rb(\hs_J) = \E\left[ \sum_{i\in \{1\to c \}} \frac{\left| \hs_J(i)-i\right|}{i^{\alpha}}\right]+ \E\left[\sum_{i \not\in \{1\to c \}} \frac{\left| \hs_J(i)-i\right|}{i^{\alpha}} \right].
 \end{align*}
 Since by Lemma \ref{lem:JordanVsHanging} and the carefully chosen coupling between Jordan and descendant ordering, vertices outside of the path
 $\{1\to c\}$ are put in the same order by the Jordan and the
 descendant ordering, then, for $i\notin \{1\to c\}$,
 $|\hs_J(i)-\hs(i)|\leq D+1$, where $D$ is the distance between $1$
 and $c$. Thus, we can control the second term of the right-hand side
 as follows:
\begin{align*}
 \E\left[\sum_{i \not\in \{1\to c \}} \frac{\left| \hs_J(i)-i\right|}{i^{\alpha}} \right] &= \E\left[\sum_{i \not\in \{1\to c \}} \frac{\left| \hs'(i)-i+\hs_J(i)-\hs'(i)\right|}{i^{\alpha}} \right] \\
& \leq \E\left[D\right]\sum_{i=1}^n\frac{1}{i^{\alpha}} + \E\left[\sum_{i \not\in \{1\to c \}} \frac{\left| \hs'(i)-i\right|}{i^{\alpha}} \right].
\end{align*}
Since $D$ is at most the arrival time of the centroid, Lemma \ref{lem:JordanVsHanging} implies that $\E[D]\leq \E[c]\leq K$.
On the other hand,
 $$ \E\left[\sum_{i\in \{1\to c \}} \frac{\left|\hs_J(i)-i\right|}{i^{\alpha}}\right]\leq \E\left[\sum_{i\in \{1\to c \}}i\right]+ \E\left[\sum_{i\in \{1\to c \}} \hs_J(i)\right]. $$
Clearly,
$$\E\left[\sum_{i\in \{1\to c \}}i\right]\leq \E[cD].$$
Since $D\leq c$ and since $c$ is dominated by an exponential random variable of mean $K$, by Lemma \ref{lem:JordanVsHanging},
$$ \E\left[ cD\right]\leq\E\left[ c^2 \right]\leq 2K^2. $$
In addition, since on the path $\{1\to c\}$, $\hs_J$ is decreasing, we have
$$ \E\left[\sum_{i\in \{1\to c \}} \hs_J(i)\right]\leq \E\left[ D\hs_J(1)\right].$$
Since $D$ and $\hs_J(1)$ are bounded by $n$, they have finite moments. Using Hölder's inequality, for any $\gamma>0$,
\begin{equation}\label{eq:Holder}
\E\left[ D\hs_J(1)\right]\leq \left( \E\left[ D^{\frac{1+\gamma}{\gamma}}\right]\right)^{\frac{\gamma}{1+\gamma}}\left( \E\left[ \hs_J(1)^{1+\gamma}\right]\right)^{\frac{1}{1+\gamma}}.
\end{equation}
Since $D\leq c$ which is dominated by an exponential random variable,
\begin{equation}\label{eq:HolderPart1}
\left( \E\left[ D^{\frac{1+\gamma}{\gamma}}\right]\right)^{\frac{\gamma}{1+\gamma}}\leq C\frac{1+\gamma}{\gamma} ,
\end{equation}
for some positive constant $C$. Next, using Lemma \ref{lem:JordanVsHanging},
$$ \PROB\left\{ \hs_J(1)\geq f(\epsilon)\right\}\leq 5\epsilon, $$
where $f(\epsilon)=2.5\frac{\log(1/\epsilon)}{\epsilon}$.  $f$ is a non-increasing function and therefore $f\left( 5\log^2(k)/k  \right)\leq k$ for all $k\geq 2$. Therefore,

$$ \PROB\left\{ \hs_J(1)\geq k\right\}\leq 25\frac{\log^2(k)}{k},\quad\text{for all}\ k\geq 2, $$
so for any $\gamma>0$,
$$ \PROB\left\{ \hs_J(1)^{1+\gamma}\geq k\right\}=\PROB\left\{ \hs_J(1)\geq k^{\frac{1}{1+\gamma}}\right\}\leq  25 \frac{\frac{1}{(1+\gamma)^2}\log^2(k)}{k^{\frac{1}{1+\gamma}}}.$$
It follows that 
\begin{align*}
\E\left[ \hs_J(1)^{1+\gamma}\right] & \leq 1+2^{1+\gamma}+ \int_{k=2}^{n^{1+\gamma}}\PROB\left\{ \hs_J(1)^{1+\gamma}\geq k\right\}\ dk\leq 1 +2^{1+\gamma}+ \int_{k=2}^{n^{1+\gamma}}\frac{\frac{25}{(1+\gamma)^2}\log^2(k)}{k^{\frac{1}{1+\gamma}}}\ dk\\
&\leq 1+2^{1+\gamma}+\frac{25}{(1+\gamma)^2}(1+\gamma)^2\log^2(n) \frac{\gamma+1}{\gamma}\left(n^{1+\gamma}\right)^{\frac{\gamma}{1+\gamma}}\\
&=1+2^{1+\gamma}+25 \frac{(1+\gamma)\log^2(n)}{\gamma}n^{\gamma}.
\end{align*}
Plugging the obtained inequality in \eqref{eq:Holder} and recalling \eqref{eq:HolderPart1}, we obtain
$$ \E\left[D\hs_J(1)\right] \leq C \frac{1+\gamma}{\gamma}\left(1+2^{1+\gamma}+25{1+\gamma\over \gamma}\log^2(n)n^{\gamma}\right). $$
Choosing $\gamma=1/\log(n)$ we get
$$ \E\left[D\hs_J(1)\right] \leq C'\log^4(n). $$
This concludes the proof of the lemma.
 \end{proof}
 
\subsection{Performance of Jordan ordering in the URRT model}\label{sec:JordanErrorurrt}
 
 In this section, we prove upper bounds for the risk $\Rb(\hs_J)$. In particular, we prove that for $\alpha\in[1,2)$, the risk $\Rb(\hs_J)$ has the same order as the optimal risk $R^*_\alpha$, defined in Section \ref{sec:urrtLowerBound}.

 \begin{theorem}\label{prop:urrtErrorUpperBound}
 Let $T\sim \text{{\sc urrt}}$. Then there exist positive constants $C$, $K$ such that for $1\leq \alpha<2$
 $$ \Rb(\hs_J)\leq K(\alpha) n^{2-\alpha}+K\sum_{i=1}^n\frac{1}{i^{\alpha}}+C\log^4(n),  $$
where $K(\alpha)=  \frac{2}{2-\alpha}+ \frac{2e^2}{(2-\alpha)^2}+\frac{2}{(2-\alpha)^3}$. Moreover, for $\alpha\geq 2$
 $$ \Rb(\hs_J)\leq C\log^4(n).  $$
 \end{theorem}

 Before proving Theorem \ref{prop:urrtErrorUpperBound}, we state a
 corollary that is a direct consequence of Theorems
 \ref{prop:urrtLowerBound} and \ref{prop:urrtErrorUpperBound}. This corollary notably implies that the Jordan ordering has a risk of optimal order for
 $\alpha\in[1,2)$. Note that one cannot hope to match the established lower bounds
for the optimal risk in a broader range of $\alpha$ for this method.
Indeed, in a uniform random recursive tree of
 size $n$, the probability that vertex $1$ is a leaf is $1/n$. Since
 leaves are ordered last by $\hs_J$, and that there are roughly $n/2$
 leaves, $\PROB\left\{ \hs_J(1)\geq n/2 \right\} \approx 1/n$. This
 implies that $\E[\hs_J(1)]\gtrsim \log(n)/2$, so $\hs_J$ has a risk of
 order at least $\log(n)$, while the lower bound is of constant order
for  $\alpha\geq 2$. We discuss in Appendix \ref{sec:RumorCentrality} the
 possibility of estimating the position of vertex $1$ better, namely
 using the rumor centroid. We conjecture this alternative method has a
 risk of optimal order for any $\alpha\geq 1$.

 \begin{cor}\label{cor:urrtErrorUpperBound}
 Let  $T\sim \text{{\sc urrt}}$. For $\alpha = 1$
 
 $$ \Rb(\hs_J) \leq \left(1+o(1)\right) 1170 R^*_1  $$
 and for $\alpha\in (1,2)$,
 
 $$ \Rb(\hs_J) \leq \left( 1+o(1)\right) \left( \frac{1}{2-\alpha}+ \frac{3}{(2-\alpha)^2}+\frac{1}{(2-\alpha)^3} \right)  70 R_{\alpha}^*.  $$
 \end{cor}

\medskip
\noindent
\emph{Proof of Theorem \ref{prop:urrtErrorUpperBound}.}
By the triangle inequality,
$$ \Rb(\hs')\leq \sum_{i=1}^n \E\left[ \frac{\hs'(i)}{i^{\alpha}} \right]+ \sum_{i=1}^n   \frac{i}{i^{\alpha}}. $$
Using the triangle inequality here is not tight and might explain why we obtain such a big gap between the lower and upper bounds. However, to get a more precise control of $|\hs'(i)-i|$, one must not only upper bound $\hs'(i)$ but prove its concentration around $i$. We did not reach this goal in this paper. Establishing concentration of $\hs'(i)$ around $i$ seems to be the natural way to make the upper bound tighter, as well as having more information about the quality of the descendant ordering (and thus Jordan ordering).
Let  $i \in [n]$ be a vertex of $T$. We first note that 
\begin{align}\label{eq:exp_sigma}
\E\left[ \hs'(i) \right] \leq \E\left[ \sum_{j=1}^n \mathbbm{1}_{\de(j)\geq \de(i)}   \right].
\end{align}
Moreover, for any $\tau_{i,j} \in \mathbb R$,
$$\PROB\left\{ \de(j)\geq \de(i)\right\} \leq \PROB\left\{ \frac{\de(j)}{n} \geq \tau_{i,j}\right\}+\PROB\left\{   \frac{\de(i)}{n} \leq \tau_{i,j} \right\}.$$
Therefore, we may upper bound \eqref{eq:exp_sigma} by
\begin{equation}\label{eq:SigmaUpperBound}
 \E\left[ \hs'(i) \right] \leq i+1+ \sum_{j> i+1} \PROB\left\{ \frac{\de(j)}{n} \geq \tau_{i,j}\right\}+\PROB\left\{  \frac{\de(i)}{n} \leq \tau_{i,j} \right\}.
\end{equation}
Let $j > i+1$ be a vertex of $T$. Note that the distributions of $\de(i)$ and $\de(j)$ in a {\sc urrt} model follow a P\'olya urn model. In particular, for any vertex $k \in [n]$,
\begin{align*}
 \PROB\left\{ k \in (T,1)_{j}  \right\}= \frac{de_{k-1}(j)+1}{k-1},
\end{align*}
and each connection of a new vertex to a descendant of $j$ is
independent of the previous ones, conditionally on $\de(j)$.  Let
$N_n:=n-j$ and let $\widetilde{W}_N=\de(j)$ be the number of
descendants of $j$ at time $n$. We thus have that
$\widetilde{W}_N=\de(j)$ follows a P\'{o}lya urn distribution, where
the P\'{o}lya urn process has balls of two colours, it is started when
$j$ is added to $T$, and it is run for a maximum number of steps
$N_n$.  From \citet[Section 3.2]{mahmoud2008polya}, we have that
$$\E \left[ \de(j)+1 \right]\ = \ \frac{n}{j},$$
and also that
\begin{align}\label{eq:LawofPolya}
\PROB\left\{ \de(j)=k \right\} \ = \ \frac{ k!  (j-1)(j)\cdots (n-k-2) }{j(j+1)\cdots (n-1)}\binom{n-j}{k}.
\end{align}
In Appendix \ref{sec:urrtAppendix}, we derive from these formulas the following upper-bounds. 
\begin{lemma}\label{lem:calcul:urrt}
$i\geq 2$ and $j>i+1$, by choosing $\tau_{i,j}  = \frac{1}{j} \log \frac{j}{i}$,
\begin{align*}
     \PROB\left\{ \frac{\de(j)}{n} \geq \tau_{i,j}\right\}+\PROB\left\{   \frac{\de(i)}{n} \leq \tau_{i,j} \right\} \leq  2e^2 e^{-   \log \frac{j}{i}  } + \frac{i}{j} \log \frac{j}{i} \leq \frac{i}{j} \left(2e^2 + \log \frac{j}{i}\right),
\end{align*}
and that for $i=1$, $j> i+1$, choosing $\tau_{1,j}=\frac{1}{j}\log(j)$,
$$ \PROB\left\{ \frac{\de(j)}{n} \geq \tau_{1,j}\right\}+\PROB\left\{   \frac{\de(1)}{n} \leq \tau_{1,j} \right\} \leq \frac{1}{j}\left(2e^2+\log(j)\right) +\frac{1}{n-1}.  $$
\end{lemma}
\noindent Once plugged into the expression of $\Rb$, for $n\geq 60$, this leads to (details in Appendix~\ref{sec:urrtAppendix})
$$ \sum_{i=1}^n \mathbb{E}\left[\frac{|\hs'(i) - i|}{i}\right]\leq 18n.$$
For $1<\alpha<2$, a similar computation yields (details in Appendix~\ref{sec:urrtAppendix})
$$  \sum_{i=1}^n \mathbb{E}\left[\frac{|\hs'(i)-i|}{i^{\alpha}}\right]  \leq \left( \frac{2}{2-\alpha}+ \frac{2e^2}{(2-\alpha)^2}+\frac{2}{(2-\alpha)^3} \right) n^{2-\alpha}.  $$
Lemma \ref{prop:ProxyIsGood} concludes the proof for $1<\alpha<2$. The proof of the case $\alpha\geq2$ lies at the end of Appendix~\ref{sec:urrtAppendix}.

\section{Preferential attachment tree}\label{sec:PA}

In this section, we consider the preferential attachment model and
investigate the performance of the Jordan ordering procedure. Since
the arguments have a similar structure to the {\sc urrt} model analyzed
in Section \ref{sec:urrt}, we omit some details of the proofs and
report them to the Appendices. Similarly to the previous section, we
first prove a minimax lower bound for the risk of any label-invariant
estimator.

\subsection{A lower bound}\label{sec:PALowerBound}

\begin{tcolorbox}
\begin{theorem}\label{prop:PALowerBound}
In the {\sc pa} model, we have, for $\alpha = 1$ and $n\geq 300$
$$ \Rb^* \geq  \frac{n^{2-\alpha}}{70}~. $$
\end{theorem}
\end{tcolorbox}
\noindent The proof is deferred to Appendix \ref{app:PALowerBound}.

\remark
In the same way as in the case of the {\sc urrt} model, we have

$$ R_{\alpha}^*\geq \frac{1}{2}~,$$
which is better than the result of Theorem \ref{prop:PALowerBound} for $\alpha>2$.

 \section{Preferential attachment tree}\label{sec:PA}

In this section, we consider the preferential attachment model and
investigate the performance of the Jordan ordering procedure. Since
the arguments have a similar structure to the {\sc urrt} model analyzed
in Section \ref{sec:urrt}, we omit some details of the proofs and
report them to the Appendices. Similarly to the previous section, we
first prove a minimax lower bound for the risk of any label-invariant
estimator.

\subsection{A lower bound}\label{sec:PALowerBound}

\begin{theorem}\label{prop:PALowerBound}
In the {\sc pa} model, we have, for $\alpha > 0$ and $n\geq 300$
$$ \Rb^* \geq  \frac{n^{2-\alpha}}{70}. $$
\end{theorem}

\noindent The proof is deferred to Appendix \ref{app:PALowerBound}.

\remark
As in the case of the {\sc urrt} model, we have

$$ R_{\alpha}^*\geq \frac{1}{2},$$
which improves on the result of Theorem \ref{prop:PALowerBound} for $\alpha>2$.

\subsection{Performance of the Jordan ordering in the PA model}

Similarly to Section \ref{sec:JordanErrorurrt}, we establish upper bounds for $\Rb(\hs_J)$. In a subsequent corollary, we bound the risk $R_\alpha(\hs_J)$ in terms of the optimal risk $\ R^*_\alpha$.

 \begin{theorem}\label{prop:PAErrorUpperBound}
 Let  $T \sim \text{{\sc PA}}$. Then, there exist positive constants $C,\ K$, such that for $\alpha\in[1,3/2)$
 $$ \Rb(\hs_J) \leq K(\alpha) n^{2-\alpha}+K\sum_{i=1}^n\frac{1}{i^{\alpha}}+ C\log^2(n)\sqrt{n} ,$$
where $K(\alpha)=\frac{2}{2-\alpha}+\frac{8\sqrt{2}+\frac{10}{\sqrt{2}}}{(3/2-\alpha)(2-\alpha)}+ \frac{20}{\sqrt{2}(2-\alpha)(3/2-\alpha)^2}  $.
 For $\alpha \geq 3/2$,
 $$ \Rb(\hs_J) \leq
 c n^{3/2} ,  $$
where $c$ is a positive constant.
 \end{theorem}

 Corollary \ref{cor:PAErrorUpperBound} is a direct consequence of
 Theorems \ref{prop:PALowerBound} and \ref{prop:PAErrorUpperBound}. It
 states that the Jordan ordering
 has a risk of optimal order for
 $\alpha\in[1,3/2)$. Let us remark that, here, the boundary value $3/2$ does
 not appear for the same reason as in the {\sc urrt} case. In the {\sc
   urrt}, the optimality result is limited to $\alpha< 2$ because of
 the error originating from the estimation of vertex $1$. Here, the
 limitation to $\alpha<3/2$ has a different origin than in the URRT model. Indeed, the descendant ordering is only order optimal for $\alpha<3/2$, meaning that even if the position of vertex
 $1$ was known, ordering vertices by the number of their descendants would not
result in a risk bound that matches the lower bound for $\alpha\geq 3/2$.

 \begin{cor}\label{cor:PAErrorUpperBound}
 Let  $T \sim \text{{\sc PA}}$. For $\alpha\in[1,3/2)$
 $$ \Rb(\hs_J) \leq \left(1+o(1)\right)  70 \left(\frac{2}{2-\alpha}+\frac{8\sqrt{2}+\frac{10}{\sqrt{2}}}{(3/2-\alpha)(2-\alpha)}+ \frac{20}{\sqrt{2}(2-\alpha)(3/2-\alpha)^2}  \right)  \Rb^* .$$
 \end{cor}

The proof of Theorem \ref{prop:PAErrorUpperBound} is reported to Appendix \ref{app:PAErrorUpperBound}.

\section{Simulations}\label{sec:Simulations}

In this section, we first report a numerical illustration of our
theoretical results on trees generated from the {\sc urrt} and {\sc
  pa} models. Then, we compare the performance of the descendant ordering
to other ordering procedures. For computational reasons, we display
results for the descendant ordering procedure. 
The descendant ordering can be computed in time
$\mathcal{O} (n\log(n))$. Also, one can find the Jordan centroid 
in linear time. 


Note that the bounds of
Lemmas 3 and 11 show that the risk of  the
descendant ordering is a good approximation of the risk of Jordan ordering.

In the first experiment, we compute the risk $\Rb(\hs')$
(see (2)) of the descendant ordering and display the
theoretical upper bound and minimax lower bound from Theorems
1 and 4
(Theorems 7 and
8 in the {\sc pa} model).

\begin{figure}
\begin{center}
\includegraphics[scale=0.5]{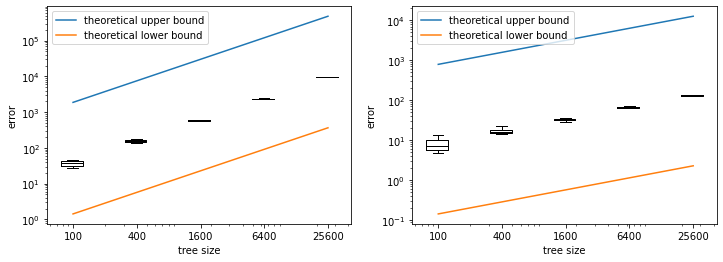}
\caption{Risk $\Rb$ of the descendant ordering versus the tree size $n$ in
  logarithmic scales, for $\alpha=1$ (left panel) and for $\alpha=1.5$
  (right panel), and for trees simulated from the {\sc urrt} model.
  Here, we sample $10$ trees for each
  size, and report a boxplot with the median, first, and last
  quartiles, for each tree size.}

\label{fig:Errorurrt}
\end{center}
\end{figure}

\begin{figure}
\begin{center}
\includegraphics[scale=0.5]{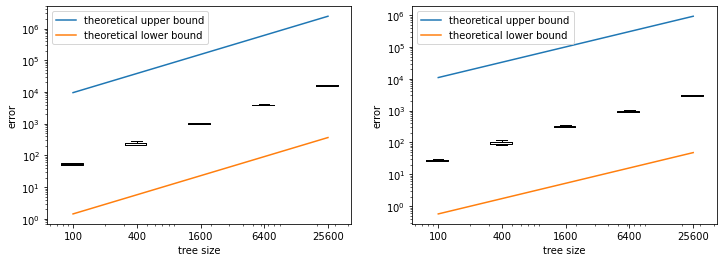}
\caption{ Risk $\Rb$  of the descendant ordering versus the  tree size
  $n$ in logarithmic scales, for $\alpha=1$ (left panel) and for
  $\alpha=1.2$ (right panel), and for trees simulated from the {\sc
    pa} model.
  Here, we sample $10$ trees for each size, and report a boxplot with
  the median, first, and last quartiles, for each tree size.}
\label{fig:ErrorPA}
\end{center}
\end{figure}

In the second experiment, we perform an empirical comparison of descendant
ordering with the three following ordering methods:
\begin{itemize}
\item \textbf{Degree} ordering, which orders the vertices by decreasing
  degree. Again, we break ties at random. Degree ordering is justified
  by the fact that the
  lower the rank of a vertex, the higher its expected degree is. Note,
  however, that ordering vertices by degree does not necessarily
  produce a recursive ordering.
\item \textbf{Spectral} method by
  \citet{recanati2018reconstructing}. This method is widely used in
   seriation problems and consists of finding the eigenvector
  associated to the second smallest eigenvalue of the Laplacian of the
  graph. Then, considering the entries of this eigenvector as a score
  function, the estimated ordering is derived by sorting these entries
  by increasing values.
\item \textbf{Reverse DMC} algorithm, proposed by
  \citet{navlkha2011}. We only compute this ordering in the PA tree, since its definition relies on a score based on a likelihood in the PA model. This algorithm is analogous to a pruning  method, which consists of ordering the vertices by sequentially
  removing all leaves from the tree and ordering the leaves removed at
  each step. In Reverse DMC, a score is computed for each leaf and the
  algorithm sequentially removes the leaf with the highest score. This
  score function corresponds to the likelihood, in the PA tree, of the leaf being the
  last vertex in the current tree, therefore, at each step, the leaf
  which is the most likely to be the last vertex arrived in the tree
  is removed.
\end{itemize}


\begin{remark}
  We note that the spectral method is a reasonable method to compare
  with in our setting of recursive trees since
  (i) spectral methods recovers the order of a Robinson matrix  \citet{recanati2018reconstructing}, and (ii)
   in the {\sc urrt} and
  {\sc pa} models, the expected value of the adjacency matrix is a
  Robinson matrix.
\end{remark}

\begin{figure}
\begin{center}
\includegraphics[scale=0.6]{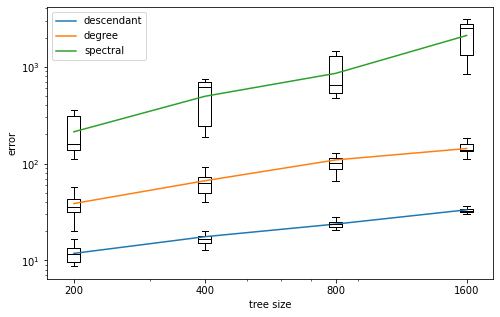}
\caption{Risk $\Rb$  versus the  tree size $n$ in logarithmic scales,
  for $\alpha=1.5$, and for trees simulated from the {\sc urrt}
  model. Here, we sample $10$ trees for each size. We compare the risk
  of descendant (blue), degree (orange), and spectral methods (green),
  and report a boxplot with the median, first, and last quartiles, for
  each tree size.
  In all settings, the descendant ordering  largely outperforms the other methods.}
\label{fig:ErrorComparisonurrt}
\end{center}
\end{figure}

\begin{figure}
\begin{center}
\includegraphics[scale=0.6]{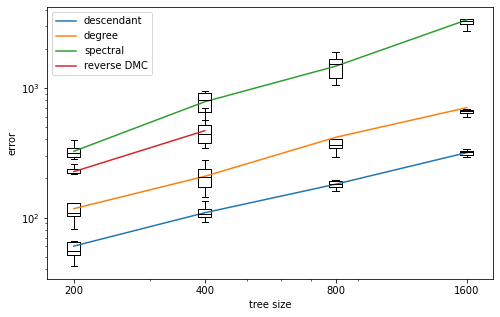}
\caption{Risk $\Rb$ versus the tree size $n$ in logarithmic scales,
  for $\alpha=1.2$, and for trees simulated from the {\sc pa}
  model. Here, we sample $10$ trees for each size, only considering
  small trees for the reverse DMC method due to its high
  computational cost. We compare the risk of descendant (blue), degree
  (orange), spectral (green), and reverse DMC (red) methods, and
  report a boxplot with the median, first, and last quartiles, for
  each tree size.
  Just like in the {\sc urrt} model, the descendant
  ordering outperforms the other methods.}
\label{fig:ErrorComparisonPA}
\end{center}
\end{figure}

Similarly to the previous experiment, we compute the risk $\Rb$ for
the four methods, on trees simulated from the {\sc urrt} or {\sc pa}
models, in multiple settings. From Figures
\ref{fig:ErrorComparisonurrt} and \ref{fig:ErrorComparisonPA}
we see that the descendant estimator has the lowest risk, for all values
of the trees sizes, and that the degree method is the second best
one. In fact, it is not surprising that the degree method performs
well for {\sc pa} trees, since, in this model, the degree has a power
law distribution and the order by degree correlates well with the
arrival times of the vertices. However, this results is more surprising for
the {\sc urrt} model, where degree-centrality is known to be sub-optimal in
the root-finding problem, see \citet{BuDeLu17}. This is discussed
further in Appendix C. Moreover, the spectral method 
has the poorest performance in both models.  A possible explanation for this is
that the random fluctuations of the adjacency matrix in the
considered recursive tree models are large, leading to a large
difference between the expected and empirical adjacency matrices in
spectral norm. This absence of concentration, which is generally
required in spectral ordering methods, could explain why this method
poorly performs in our setting.  Finally, the Reverse DMC algorithm
performs similarly poorly as the spectral method in the {\sc pa} model.

\newpage

\appendix

\section{Appendix}

In this appendix we discuss some issues concerning the choice of the
parameter $\alpha$ in the definition of the risk and ordering
according to degrees. Some elements of the proofs for the {\sc urrt}
model are also reported here. The final sections contain the proofs
of all results in the {\sc pa} model.

\subsection{A remark on the choice of the risk}\label{app:alpha}

The risk $R_\alpha$ defined in \eqref{eq:risk2} leads to a meaningful
performance measure in the {\sc urrt} and {\sc pa} models only for
some values of $\alpha$. In particular, for $\alpha < 1$, it is easy
to see that the risk of a random permutation is of the same order as 
the established lower bound, both in the {\sc urrt}
and {\sc pa} models. More precisely, let $\Sigma$ be a permutation
chosen uniformly at random. Simple computation for $\alpha <1$
leads to
$$\Rb(\Sigma)\leq c_{\alpha}n^{2-\alpha},$$
for some positive constant $c_{\alpha}$. On the other hand, 
Theorems \ref{prop:urrtLowerBound} and \ref{prop:PALowerBound} imply that
for $\alpha <1$, $R_{\alpha}^*\geq c'_{\alpha}n^{2-\alpha}$.
Therefore, with $\alpha <1$, the random ordering has
a risk of the same order as that of the optimal one.
This is why we restrict our analysis of the
risk to $\alpha \geq 1$. Our analysis of the Jordan
ordering proves that this method has a risk of optimal order for
$\alpha\in[1,2)$ in the {\sc urrt} case and $\alpha\in[1,3/2)$ in the
{\sc pa} tree.

It is also possible to change the risk by replacing
  $1/\sigma(i)^{\alpha}$ by $f(\sigma(i))$, for some non-increasing
  function $f$. An example is $f(t)=\IND{t\leq \tau_n}$, for some
  threshold $\tau_n$. Such a risk measures the quality of the ordering
  on the first $\tau_n$ vertices only. For many choices of $f$, our
  analysis can be directly adapted
  However, for rapidly decreasing functions (such as the step function
  above), the Jordan ordering is not optimal up to a constant
  factor.
  This suggests that one should investigate methods that are better at ordering the early vertices.

\subsection{A remark on rumor centrality}\label{sec:RumorCentrality}

We conjecture that in the {\sc urrt} model, there exists an ordering
procedure whose risk is of the order of $n^{2-\alpha}$
for any $\alpha\geq 1$, matching that of the minimax lower bound. Indeed,
in our analysis, the risk is decomposed in two parts. First, a part
coming from the difference between the Jordan and the descendant
ordering (i.e, the error made by estimating the position of vertex $1$
by the Jordan centre), second the risk of the descendant ordering. A possible
way to improve our bound on the risk is to estimate the position of
vertex $1$ more precisely. To do so, using the rumor centrality
appears to be a promising option. Indeed, due to recent results
from \citet{CrXu21}, in the {\sc urrt} model, the
rumor centrality orders vertices by their likelihood of being vertex
$1$. In particular, using the rumor centrality is optimal for
minimizing the size of a conficence set containing the root, outperforming Jordan centrality
(\citet{BuDeLu17}). However,  one step in the analysis is missing. Copying the proof
of Lemma \ref{prop:ProxyIsGood}, with the rumor center instead of the
Jordan center, one needs to bound the moment of order $1+\gamma$ of the arrival
time of the rumor centre. Bounding it by a
constant (for any value of $\gamma$) would be sufficient to prove that this new
ordering procedure has a risk of optimal order for any $\alpha\geq 1$.

\subsection{A remark on ordering by degree}\label{sec:degree}

\begin{figure}
\begin{center}
\includegraphics[scale=0.55]{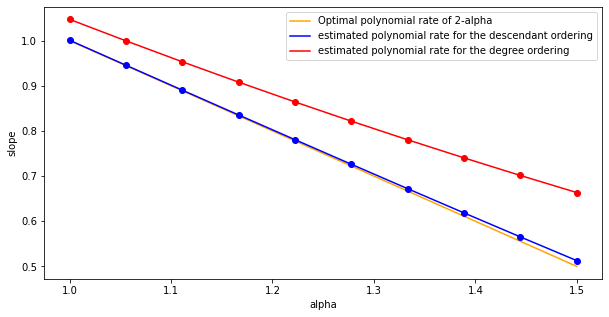}
\caption{An estimation of the rate at which the risk increases with
  the size of the tree for different values of $\alpha$ in the {\sc
    pa} tree. Here, we compare the case of the descendant and 
  degree ordering. Proposition 7 shows that the optimal risk
  grows as $n^{2-\alpha}$,
  and that the the risk of  descendant ordering grows at the same rate (see
  Proposition 8). This experiment
  confirms these results. For the degree ordering, this plot
  suggests that the risk grows faster than $n^{2-\alpha}$.
  }
\label{fig:PADegreeRiskRate}
\end{center}
\end{figure}

As discussed in the above simulations, a simple ordering procedure is by the
degrees. Simulations suggest
that it does not perform as well as Jordan ordering, and it may produces non-recursive ordering. Nonetheless,
it is a simple procedure worth mentioning. Since in a {\sc pa} tree
 the degree of a given node follows a P\'{o}lya urn distribution,
analysing the performance of the degree ordering is similar to the
analysis carried out for the Jordan ordering. However, the simulations results displayed in Figure
\ref{fig:PADegreeRiskRate} suggest that for any $\alpha\in[1,3/2)$ the
risk of the degree ordering grows at a faster rate than $n^{2-\alpha}$. Both in the URRT and PA model, for sizes of trees $\{1000,2000,4000,8000\}$, we sample $10$ trees at each size and compute the risk of the descendant and degree ordering for different values of $\alpha$. Then, for each value of $\alpha$, we perform a linear regression on the log-plot of the risk, to estimate the exponent of the polynomial.

\begin{figure}
\begin{center}
\includegraphics[scale=0.55]{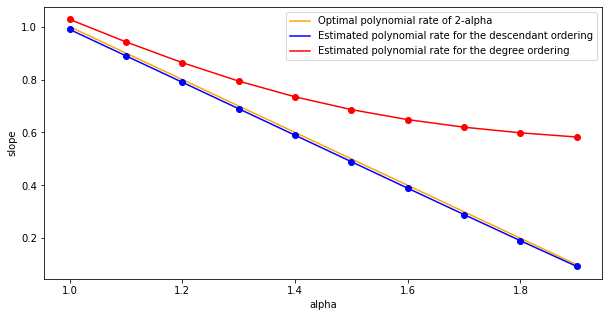}
\caption{Estimation of the rate at which the risk increases with the
  size of the tree for different values of $\alpha$ in the {\sc
    urrt}. Here we compare the case of descendant and degree
  ordering. Proposition 1 shows that the optimal risk grows
  as $n^{2-\alpha}$, and
  that the Jordan and descendant orderings' risks grow at the same rate
  (see Proposition 4). This experiment is
  in accordance with these results. For the degree ordering, this plot
  suggests that the risk grows at a rate faster than $n^{2-\alpha}$.}
\label{fig:urrtDegreeRiskRate}
\end{center}
\end{figure}

On the other hand, ordering vertices by their degree in the {\sc urrt}
is known to be suboptimal for finding the root, as there are many
vertices with much higher degree (\cite{Esl22}). Simulation
results displayed in Figure \ref{fig:urrtDegreeRiskRate} suggest that,
for most values of $\alpha$, 
the risk of ordering by degree in the {\sc urrt} model grows at a faster rate than
$n^{2-\alpha}$ for any $\alpha\in (1,2)$. On the other hand, observing Figure
\ref{fig:urrtDegreeRiskRate}, it seems like, for $\alpha=1$, the
degree ordering may have a risk growing at the optimal rate of
$n$.

In the Supplementary Material, where the empirical performance of
different ordering procedures are compared, the degree ordering is the
method with second best performance. The above simulations suggest that
all the other tested methods have risks growing at a faster rate than
$n^{2-\alpha}$.

\subsection{Proof of Theorem \ref{prop:urrtErrorUpperBound} }
\label{sec:urrtAppendix}

Here, we present the arguments to complete the proof of Theorem
\ref{prop:urrtErrorUpperBound}. We recall that we need to upper bound
$ \sum_{i=1}^n \mathbb{E}\left[\frac{|\hs'(i)-i|}{i^{\alpha}}\right]$,
and that we reduced in (\ref{eq:SigmaUpperBound}) the problem to upper bounding
$\PROB\left\{ \frac{\de(j)}{n} \geq \tau_{i,j}\right\}+\PROB\left\{
  \frac{\de(i)}{n} \leq \tau_{i,j} \right\}$, which is done in Lemma~\ref{lem:calcul:urrt}. 
  
 \begin{proof}[Proof of Lemma~\ref{lem:calcul:urrt}]
 From
\eqref{eq:LawofPolya}, we have
$$\PROB\left\{ \de(j)=k \right\} \ = \ \frac{ k!  (j-1)(j)\cdots (n-k-2) }{j(j+1)\cdots (n-1)}\binom{n-j}{k}.$$
Re-arranging the factors, 
\begin{align*}
\PROB\left\{ \de(j)=k \right\} \ &= \ k!  \frac{ (n-k-2)! }{(j-2)!} \frac{(j-1)!}{(n-1)!} \frac{(n-j)!}{k!(n-j-k)!} \\
&= (j-1) \frac{(n-k-2)!}{(n-j-k)!} \frac{(n-j)!}{(n-1)!}
\end{align*}
Since $j\geq3$,
\begin{align*}
\PROB\left\{ \de(j)=k \right\} \ &= (j-1)  \frac{(n-j-k+1)\cdots(n-k-2)}{(n-j+1)\cdots (n-1)}\\
&= \frac{j-1}{n-1} \frac{n-j-k+1}{n-j+1}\cdots \frac{n-k-2}{n-2}~
\\
& \leq \ \frac{j-1}{n-1} \left( 1-\frac{k}{n-2}\right)^{j-2} .
\end{align*}
Therefore, for $n, j \geq 3 $, we can upper bound the second term of \eqref{eq:SigmaUpperBound} by
\begin{align*}
\PROB\left\{ \frac{\de(j)}{n} \geq \tau_{i,j}\right\} \ & \leq \  \sum_{(\tau_{i,j}n)\leq k \leq n} \frac{j-1}{n-1} \left( 1-\frac{k}{n}\right)^{j-2} \\
& \leq \frac{(j-1)n}{n-1}  \int_{\tau-\frac{1}{n}}^1 (1-t)^{j-2}dt=\frac{n}{n-1} \left( 1-\tau_{i,j}+\frac{1}{n}\right)^{j-1}  \leq 2e^2 e^{- j  \tau_{i,j} },
\end{align*}
using $\log(1+x) \leq x$. Moreover, for $i\geq2$, we bound the third term of \eqref{eq:SigmaUpperBound} by
\begin{align*}
     \mathbb{P}\left\{\frac{\de(i)}{n} \leq \tau_{i,j} \right\} &\leq \sum_{k \in [1,\tau_{i,j} n]}  \frac{i-1}{n-1}\left(1-\frac{k}{n-2}\right)^{i-2} \\
     &\leq \tau_{i,j}  i \left(1-\frac{1}{n-2} \right)^{i-2} \leq \tau_{i,j}  i.
\end{align*}
Since for $i=1$, $\PROB\left\{ \de(1)=k\right\}=1/(n-1)$,  
$$ \PROB \left\{\frac{\de(1)}{n} \leq \tau_{i,j} \right\} \leq \sum_{k \in [1,\tau_{i,j} n]}  \frac{1}{n-1}\leq \tau_{i,j}+\frac{1}{n-1}. $$
Therefore, for $i\geq 2$ and $j>i+1$, by choosing $\tau_{i,j}  = \frac{1}{j} \log \frac{j}{i}$, we obtain that
\begin{align*}
     \PROB\left\{ \frac{\de(j)}{n} \geq \tau_{i,j}\right\}+\PROB\left\{   \frac{\de(i)}{n} \leq \tau_{i,j} \right\} \leq  2e^2 e^{-   \log \frac{j}{i}  } + \frac{i}{j} \log \frac{j}{i} \leq \frac{i}{j} \left(2e^2 + \log \frac{j}{i}\right),
\end{align*}
and for $i=1$, $j> i+1$, choosing $\tau_{1,j}=\frac{1}{j}\log(j)$ we obtain that 
$$ \PROB\left\{ \frac{\de(j)}{n} \geq \tau_{1,j}\right\}+\PROB\left\{   \frac{\de(1)}{n} \leq \tau_{1,j} \right\} \leq \frac{1}{j}\left(2e^2+\log(j)\right) +\frac{1}{n-1}.  $$
 \end{proof}

\noindent Plugging the upper-bounds of Lemma~\ref{lem:calcul:urrt} in  (\ref{eq:SigmaUpperBound}), for $\alpha=1$, we get
\begin{align}\label{eq:ProbaUpperBound}
   \lefteqn{ \sum_{i=1}^n \mathbb{E}\left[\frac{|\hs'(i) - i|}{i}\right]  \leq n+\sum_{i=1}^n \mathbb{E}\left[\frac{\hs'(i)}{i}\right]} \nonumber \\
    &\leq n + \sum_{i=2}^n \frac{1}{i} \left( i+1+  \sum_{j=i+2}^n \frac{i}{j} \left(2e^2 + \log \frac{j}{i}\right) \right) + 2+\sum_{j=3}^n \left( \frac{1}{j}\left(2e^2+\log(j)\right) +\frac{1}{n-1} \right) \nonumber \\
    &\leq 2 n+3+\log(n) +  \sum_{j=3}^n \frac{1}{j} \sum_{i=1}^{j-2}  \left(2e^2 + \log \frac{j}{i}\right).
\end{align}
Since 
$$ \sum_{i=1}^{j-2}  \log \frac{j}{i}\leq \log\left( \frac{j^j}{j!}\right), $$
and that the Stirling formula implies that $j!\geq (j/3)^j$,
$$ \sum_{i=1}^{j-2}  \log \frac{j}{i}\leq j \log(3). $$
Plugging this in \eqref{eq:ProbaUpperBound} yields
$$ \sum_{i=1}^n \mathbb{E}\left[\frac{|\hs'(i) - i|}{i}\right] \leq 3+\log(n)+ (2+2e^2+\log   3 )n, $$
which in turn proves that for $n\geq 60$,
$$ \sum_{i=1}^n \mathbb{E}\left[\frac{|\hs'(i) - i|}{i}\right]\leq 18n.$$
For $1<\alpha<2$, a similar calculation yields
\begin{align*}
    \sum_{i=1}^n \mathbb{E}\left[\frac{|\hs'(i)-i|}{i^{\alpha}}\right] &\leq \frac{1}{2-\alpha}n^{2-\alpha}+\sum_{i=1}^n \mathbb{E}\left[\frac{\hs'(i)}{i^{\alpha}}\right]  \\ 
    &\leq \frac{1}{2-\alpha}n^{2-\alpha}+1+  \sum_{i=1}^n \frac{1}{i^{\alpha}} \left( i+1 \sum_{j=i+2}^n \frac{i}{j} \left(2e^2 + \log \frac{j}{i}\right) \right) \\
    &\leq \frac{2}{2-\alpha}n^{2-\alpha}+1+\zeta(\alpha) +\sum_{j=3}^n  \frac{1}{j}\sum_{i=1}^{j-2} \frac{2e^2}{i^{\alpha-1}}+\frac{1}{i^{\alpha-1}}\log \frac{j}{i}    \\
    &\leq  \frac{2}{2-\alpha}n^{2-\alpha}+1+\zeta(\alpha)+\frac{2e^2}{(2-\alpha)^2}n^{2-\alpha}+\sum_{j=3}^n  \frac{1}{j}\sum_{i=1}^{j-2} \frac{1}{i^{\alpha-1}}\log \frac{j}{i} .
\end{align*}
Recall that $\zeta$ denotes the Riemann zeta function.
We may upper bound
$$ \sum_{i=1}^{j-2} \frac{1}{i^{\alpha-1}}\log \frac{j}{i}\leq 2 \int_1^j \frac{1}{t^{\alpha-1}}\log\left(\frac{j}{t}\right), $$
which in turn can be evaluated by integration by parts, leading to
$$ \sum_{i=1}^{j-2} \frac{1}{i^{\alpha-1}}\log \frac{j}{i}\leq
\frac{2}{(2-\alpha)^2}j^{2-\alpha}.$$
Finally
$$  \sum_{i=1}^n \mathbb{E}\left[\frac{|\hs'(i)-i|}{i^{\alpha}}\right]  \leq \left( \frac{2}{2-\alpha}+ \frac{2e^2}{(2-\alpha)^2}+\frac{2}{(2-\alpha)^3} \right) n^{2-\alpha}.$$
For $\alpha\geq 2$, we similarly get

\begin{equation}\label{eq:appendixProofalphaGeq2}
 \sum_{i=1}^n \mathbb{E}\left[\frac{|\hs'(i)-i|}{i^{\alpha}}\right]  \leq C,
\end{equation}
for some positive constant $C$.

\subsection{Proof of the minimax lower bound in the PA model}\label{app:PALowerBound}

Here we prove Theorem \ref{prop:PALowerBound}

\begin{proof}
The proof follows the same argument as that of Theorem
\ref{prop:urrtLowerBound}. It suffices to check that the event 
$$ \Omega_j:=\left\{ \tau(j) \text{ and } \tau(\left\lfloor n/4\right\rfloor +j) \text{ are leaves, connected to vertices of rank in } [n/2] \right\}$$
has a probability bounded away from $0$. Proceeding as in the proof of Theorem \ref{prop:urrtLowerBound}, we get
\begin{align*}
\PROB\left\{ \Omega_j \right\}  = & \frac{2\left(\left\lfloor n/2 \right\rfloor-1 \right)}{2(j-2)}\prod_{k=j+1}^{\left\lfloor n/4\right\rfloor +j-1}\frac{2k-3}{2(k-1)} \frac{2\left(\left\lfloor n/2 \right\rfloor -1 \right)}{2\left(\left\lfloor n/4\right\rfloor +j-2 \right)}\prod_{k=\left\lfloor n/4 \right\rfloor +j+1}^n \frac{2k-4}{2(k-1)}\\
= &  \frac{\left(\left\lfloor n/2 \right\rfloor-1 \right)}{(j-2)}\frac{2j-1}{2\left( \left\lfloor n/4 \right\rfloor +j-2 \right)}  \frac{\left(\left\lfloor n/2 \right\rfloor -1 \right)}{\left(\left\lfloor n/4\right\rfloor +j-2 \right)}\frac{\left(\left\lfloor n/4\right\rfloor +j-1 \right)\left(\left\lfloor n/4\right\rfloor +j \right)}{(n-1)(n-2)}\\
=& \frac{\left(\left\lfloor n/2 \right\rfloor -1 \right)^2}{(n-1)(n-2)}\frac{2j-1}{2j-4}\frac{\left(\left\lfloor n/4\right\rfloor +j-1 \right)\left(\left\lfloor n/4\right\rfloor +j \right)}{\left( \left\lfloor n/4 \right\rfloor +j-2 \right)^2}\\
\geq & \frac{1}{4}\left(1- \frac{5}{n}\right)^5.
\end{align*}
\end{proof}

\subsection{Descendant ordering in the PA model}\label{app:proxPA}

Here, we analyze the descendant ordering in the {\sc pa} model.
Recall the notation introduced in Section \ref{sec:proxyurrt}: the
centrality measure
$\psi'(u)=n+1-\de(u)$, and the corresponding ordering $\hs'$.
In the next lemma we prove that $\psi'$ and
$\psi$ coincide for most vertices and provide a control both on the
number of vertices for which they differ and the estimated arrival
time of vertex $1$.

\begin{lemma}\label{lem:JordanVsHangingPA}
Let $c$ be the rank of a Jordan's centroid, and let $\{1\to c\}$ be the set of vertices on the path from the root to the centroid. Then
\begin{itemize}
\item $ \forall v \in [n] \backslash \{1\to c \}, \ \psi_T(v)=\psi'_T(v); $
\item there exists an universal constant $K$ such that $c$ is
  stochastically dominated by an exponential random variable with
  parameter $K$; 
\item for any $\epsilon>0$, with probability at least $1-\epsilon$
$$\hs_J(1)\leq\frac{C}{\epsilon^2}\exp\left(\sqrt{C\log\left(\frac{1}{\epsilon}\right)} \right).$$
\end{itemize}
\end{lemma}

\begin{proof} The first part of the proof is identical to the proof of
  Lemma \ref{lem:JordanVsHanging}.  First, we use Theorem 6 of 
  \cite{wagner2019centroid}, which extend the result of 
  \citet{moon2002centroid} from uniform random recursive trees to
  preferential attachment trees. Using their result, we obtain that
$$ \PROB\left\{ c\geq k \right\} \leq \sum_{j=k}^{\infty} \frac{(-\log(2)/2)^j}{j!}, $$
so there exists an exponential random variable of parameter $K$ such that $c\leq \mathcal{E}(K)$. Using Corollary 3.3.b of \citet{BaBh20}, we have that the event
$$ \hs_J(1) \leq \frac{C}{\epsilon^2}\exp\left(\sqrt{C\log\left(\frac{1}{\epsilon}\right)} \right), $$
holds with probability at least $1-\epsilon$. This concludes the proof of the lemma.
\end{proof}

The next lemma allows us to compare the risk of Jordan and descendant
ordering.

\begin{lemma}\label{prop:ProxyIsGoodPA}
Let  $T \sim \text{PA}$. Then, there exist positive constants $C, \ K$, such that, for $\alpha>0$
$$ \Rb(\hs_J) \leq \Rb(\hs') +K\sum_{i=1}^n\frac{1}{i^{\alpha}}+C\log^2(n)\sqrt{n}. $$
\end{lemma}

\begin{proof}
The proof is similar to the one of Lemma \ref{prop:ProxyIsGood}, using the same coupling between the Jordan and descendant ordering. Recalling that $D$ is the distance between vertices $1$ and $c$, we have
\begin{align*} 
\E\left[\sum_{i\in \{1\to c \}} \frac{\left|\hs_J(i)-i\right|}{i^{\alpha}}\right] & \leq \E\left[\sum_{i\in \{1\to c \}}i\right]+ \E\left[\sum_{i\in \{1\to c \}} \hs_J(i)\right]\\
&\\
&\leq \frac{1}{2}\E\left[D^2\right]+\E\left[D\hs_J(1)\right] . 
\end{align*}
As in Lemma \ref{prop:ProxyIsGood}, we use the fact that $D\leq c$ and the domination of $c$ by an exponential random variable (see Lemma \ref{lem:JordanVsHangingPA}) to get that
$$ \frac{1}{2}\E\left[ D^2 \right]\leq K^2. $$
Then, it follows from Hölder's inequality that
\begin{equation}\label{eq:HolderPA}
\E\left[ D\hs_J(1)\right]\leq \left( \E\left[ D^{\frac{1+\gamma}{\gamma}}\right]\right)^{\frac{\gamma}{1+\gamma}}\left( \E\left[ \hs_J(1)^{1+\gamma}\right]\right)^{\frac{1}{1+\gamma}}.
\end{equation}
Using once again the domination of $D$ by an exponential random variable,
$$\left( \E\left[ D^{\frac{1+\gamma}{\gamma}}\right]\right)^{\frac{\gamma}{1+\gamma}}\leq C\frac{1}{1+\gamma}{\gamma},$$
for some positive constant $C$.  Next, using Lemma \ref{lem:JordanVsHangingPA},
$$ \PROB\left\{ \hs_J(1)\geq f(\epsilon)\right\}\leq \epsilon, $$
where
$f(\epsilon)=\frac{C}{\epsilon^2}\exp\left(\sqrt{C\log\left(\frac{1}{\epsilon}\right)}
\right)$. The function $f$ is a non-increasing, therefore $f\left( \frac{C}{\sqrt{k}}\exp\left(\sqrt{C\log(k)}\right)  \right)\leq k$. So
$$ \PROB\left\{ \hs_J(1)\geq k\right\}\leq  \frac{C}{\sqrt{k}}\exp\left(\sqrt{C\log(k)}\right) . $$
Following the same steps as in Lemma \ref{prop:ProxyIsGood}, and choosing $\gamma=1/\log(n)$, yields
$$ \E\left[D\hs_J(1)\right] \leq C\log^2(n)\sqrt{n}, $$
which concludes the proof of the lemma.
\end{proof}

\subsection{Performance of Jordan ordering in the PA model}\label{app:PAErrorUpperBound}

In this section we prove Theorem \ref{prop:PAErrorUpperBound}.

\begin{proof}
  Similarly to the {\sc urrt} case, in the {\sc pa} model, the number
  of descendants of a vertex is distributed as a P\'{o}lya urn. This
  well-know fact is easily seen
  since in the {\sc pa} model, sampling a vertex with a probability
  proportional to its degree is the same as sampling an edge uniformly
  at random and picking one of its endpoints at random. In turn, it is
  the same as picking a half edge uniformly at random. Therefore, the
  resulting P\'{o}lya urn has slightly different initial conditions
  than in the {\sc urrt}. Such P\'{o}lya urns are well understood. 
   In particular, by \citet[Section 3.2]{mahmoud2008polya}, for a
  vertex $i \in [n]$, the distribution of $\de(i)$ is given by
\[
\PROB\left\{ \de(i)=k \right\}  
 =\frac{ \left(1 3  \cdots  (2k-1)\right)\left( (2i-3) (2i-1)  \cdots  (2n-2k-5)\right) }{(2i-2) 2i  \cdots  (2n-4)} \binom{n-i}{k}.
\]
Re-arranging the terms in the above expression, 
\begin{eqnarray}
  \label{eq:pa_law_descendants}
\PROB\left\{ \de(i)=k \right\}
&= & \underbrace{\frac{1 \cdot 3  \cdots  (2k-1)}{k!}}_{\defeq A} \cdot
                                 \underbrace{\frac{ (2i-3) (2i-1)
                                 \cdots  (2n-5)}{(2i-2) 2i  \cdots
                                 (2n-4)}}_{\defeq B}  \nonumber \\
&  & \cdot\underbrace{ \frac{(n-i)!/(n-i-k)!}{ (2n-2k-3) (2n-2k-1) \cdots
                                                                     (2n-5) }}_{\defeq C}.
\end{eqnarray}
We bound each term on the right-hand side of \eqref{eq:pa_law_descendants}. 
First, for $k\geq1$,
$$ A=\frac{1}{k} \prod_{j=1}^{k-1}\frac{2j+1}{j}=\frac{2^{k-1}}{k} \prod_{j=1}^{k-1}\left( 1+ \frac{1}{2j}\right).  $$
Since
$$ \prod_{j=1}^{k-1}\left( 1+ \frac{1}{2j}\right) =\exp\left( \sum_{j=1}^{k-1} \log\left( 1+\frac{1}{2j}\right) \right) \leq \exp\left( \sum_{j=1}^{k-1} \frac{1}{2j} \right) \leq \sqrt{k}, $$
then,
$$  A\leq \frac{2^{k-1}}{\sqrt{k}}. $$
Second, we have 
\begin{align*}
 B&=\prod_{j=i}^{n-1}\frac{2j-3}{2j-2}=\prod_{j=i}^{n-1}\left( 1- \frac{1}{2j-2}\right) \\
 &=\exp\left( \sum_{j=i}^{n-1} \log\left(1-\frac{1}{2j-2}\right) \right) \leq \exp\left( -\sum_{j=i}^{n-1}\frac{1}{2j-2}\right)  \leq 2\sqrt{\frac{i}{n}}.
\end{align*}
Finally, we have that
$$ C=\prod_{j=n-k-2}^{n-3}\frac{j-i-3}{2j+1}=\frac{1}{2^{k-1}}\prod_{j=n-k-2}^{n-3} \left(1-\frac{i+2.5}{j+0.5}\right)  \leq \frac{1}{2^{k-2}}\left( 1-\frac{k}{n}\right)^i.$$
Plugging these bounds into \eqref{eq:pa_law_descendants},
we get, for any $k\geq1$,
$$  \PROB\left\{ \de(i)=k \right\} \leq 4\sqrt{\frac{i}{kn}}\left(1-\frac{k}{n}\right)^i. $$
Since $\PROB\left\{de(i)=0 \right\}=(i-1)/(n-1)$, for any $\tau > 0$,

\begin{equation*}
\PROB\left\{ \frac{\de(i)}{n}\leq \tau \right\} \leq \frac{i}{n}+ \sum_{k=1}^{\lfloor n\tau\rfloor}4 \sqrt{\frac{i}{kn}} \leq\frac{i}{n}+ 4 \sqrt{i}\int_0^{\tau}\frac{1}{\sqrt{t}}dt\leq\frac{i}{n}+ 8\sqrt{i\tau}.
\end{equation*}
Since $i/n\leq 2\sqrt{i\tau}$ (because we choose $\tau\geq 1/(2n)$, see hereafter), we get 

\begin{equation}\label{eq:PAdei}
\PROB\left\{ \frac{\de(i)}{n}\leq \tau \right\} \leq10\sqrt{i\tau}.
\end{equation}
Now, for $j \in [n]$, we have

\begin{align}\label{eq:PAdej}
 \PROB\left\{ \frac{\de(j)}{n}\geq \tau \right\} &\leq \sum_{k=\lceil n\tau\rceil}^{n} 4\sqrt{\frac{j}{kn}}\left( 1-\frac{k}{n}\right)^j \leq 4\sqrt{\frac{j}{\tau}}\sum_{k=\lceil n\tau\rceil}^{n} \frac{1}{n}\left( 1-\frac{k}{n}\right)^j~ \nonumber \\
 &\leq  4\sqrt{\frac{j}{\tau}}\int_{\frac{\lceil n\tau \rceil-1}{n}}^1 (1-t)^j dt  \leq  \frac{4}{\sqrt{j\tau}}\left(1- \frac{n\tau-1}{n} \right)^j  .
\end{align}
We set $\tau=\frac{1+\log(j/i)}{2j}$. Plugging it in \eqref{eq:PAdei} leads to
$$\PROB\left\{ \frac{\de(i)}{n}\leq \tau \right\} \leq10\sqrt{\frac{i}{2j}}\sqrt{1+\log(j/i)},$$
and plugging it in \eqref{eq:PAdej} we get
\begin{align*}
\PROB\left\{ \frac{\de(j)}{n}\geq \tau \right\} & \leq 4\sqrt{2}\exp\left( j\log\left( 1-\frac{n\tau-1}{n}\right) \right)\\
&\leq \leq 4\sqrt{2}\exp\left(\frac{-\log(j/i)}{2} \right)\exp\left( -j\frac{\frac{n}{2j}-1}{n} \right)\\
&\leq 8\sqrt{2}\sqrt{\frac{i}{j}}.
\end{align*}
Summing these two bounds, we get
$$\PROB\left\{ \de(i)\leq \de(j) \right\} \leq \left(8\sqrt{2}+ \frac{10}{\sqrt{2}}\sqrt{1+\log(j/i)} \right)\sqrt{\frac{i}{j}}. $$
Following similar calculations as in Section \ref{sec:JordanErrorurrt} and Appendix \ref{sec:urrtAppendix},
\begin{align*}
 \sum_{i=1}^n \E\left[\frac{|\hs'(i)-i|}{i^{\alpha}}\right] & \leq \frac{1}{2-\alpha}n^{2-\alpha}+ \sum_{i=1}^n \frac{1}{i^{\alpha}} \left( i+\sum_{j=i+1}^n \left(8\sqrt{2}+\frac{10}{\sqrt{2}}+\frac{10}{\sqrt{2}}\log(j/i)\right)\sqrt{\frac{i}{j}} \right) \\
 & \leq \frac{2}{2-\alpha}n^{2-\alpha}+ \sum_{j=1}^n  \left( \frac{1}{\sqrt{j}} \sum_{i=1}^{j-1} \left(8\sqrt{2}+\frac{10}{\sqrt{2}}+\frac{10}{\sqrt{2}}\log(j/i)\right)i^{1/2-\alpha}  \right).
 \end{align*}
 For $\alpha\in[1,3/2)$ we obtain
 $$ \sum_{i=1}^n \E\left[\frac{|\hs'(i)-i|}{i^{\alpha}}\right]  \leq \left(\frac{2}{2-\alpha}+\frac{8\sqrt{2}+\frac{10}{\sqrt{2}}}{(3/2-\alpha)(2-\alpha)}+ \frac{20}{\sqrt{2}(2-\alpha)(3/2-\alpha)^2}  \right)n^{2-\alpha},$$
while for $\alpha\geq 3/2$
$$ \sum_{i=1}^n \E\left[\frac{|\hs'(i)-i|}{i^{\alpha}}\right] \leq C n^{3/2} ,$$
for some positive constant $C$. This concludes the proof of Theorem \ref{prop:PAErrorUpperBound}.

\end{proof}

\bibliographystyle{plainnat}
\bibliography{bibli}

\end{document}